\documentclass{article}

\PassOptionsToPackage{numbers, sort}{natbib}  %

\usepackage[preprint]{neurips_2022}

\usepackage{mathtools}
\usepackage{xspace}
\usepackage{amsmath}
\usepackage{bbm}
\usepackage{amssymb}
\usepackage{pifont}
\usepackage{caption}
\usepackage{subcaption}
\usepackage{graphicx}
\usepackage{wrapfig}
\usepackage{enumitem}
\usepackage{multirow}
\usepackage{comment}

\usepackage[utf8]{inputenc} %
\usepackage[T1]{fontenc}    %
\usepackage{hyperref}       %
\usepackage{url}            %
\usepackage{booktabs}       %
\usepackage{amsfonts}       %
\usepackage{nicefrac}       %
\usepackage{microtype}      %
\usepackage{xcolor}         %
\usepackage{amsthm}         %
\usepackage{hyperref}

\newcommand{\ptilde}{\Tilde{p}}

\newcommand{\phat}{\hat{p}}
\newcommand{\rhat}{\hat{r}}

\newcommand{\btheta}{\boldsymbol{\theta}}
\newcommand{\bTheta}{\boldsymbol{\Theta}}

\newcommand{\bw}{\boldsymbol{w}}
\newcommand{\bx}{\boldsymbol{x}}
\newcommand{\bxo}{\boldsymbol{x}_{o}}

\newcommand{\bz}{\boldsymbol{z}}

\newcommand{\joint}{K}
\newcommand{\marginal}{0}

\newcommand{\identity}{\boldsymbol{I}}

\newcommand{\Mid}{\, \Vert \,}
\renewcommand{\mid}{\, | \,}

\DeclareMathOperator*{\argmin}{arg\,min}

\newcommand{\REJABC}{\textsc{rej-abc}\xspace}
\newcommand{\NLE}{\textsc{nle}\xspace}
\newcommand{\NPE}{\textsc{npe}\xspace}
\newcommand{\NRE}{\textsc{nre}\xspace}

\newcommand{\NREA}{\textsc{nre-a}\xspace}
\newcommand{\NREB}{\textsc{nre-b}\xspace}
\newcommand{\NREC}{\textsc{nre-c}\xspace}
\newcommand{\SMCABC}{\textsc{smc-abc}\xspace}
\newcommand{\SNLE}{\textsc{snle}\xspace}
\newcommand{\SNPE}{\textsc{snpe}\xspace}
\newcommand{\SNRE}{\textsc{snre}\xspace}

\newcommand{\SNREB}{\textsc{snre-b}\xspace}

\newcommand{\MCMC}{\textsc{mcmc}\xspace}

\newcommand{\SBI}{\textsc{sbi}\xspace}
\newcommand{\KLD}{\textsc{kld}\xspace}
\newcommand{\ROC}{\textsc{roc}\xspace}
\newcommand{\AUC}{\textsc{auc}\xspace}

\newtheorem{Lem}{Lemma}

\everypar{\looseness=-1}

\title{Contrastive Neural Ratio Estimation\\ for Simulation-based Inference}
\author{
	Benjamin Kurt Miller \\
	University of Amsterdam \\ 
	\texttt{b.k.miller@uva.nl}
	\And
	Christoph Weniger \\
	University of Amsterdam \\
	\texttt{c.weniger@uva.nl}
	\And
	Patrick Forr\'e \\
	University of Amsterdam \\
	\texttt{p.d.forre@uva.nl}
}
\date{May 2022}

\begin{document}

\maketitle

\begin{abstract}
Likelihood-to-evidence ratio estimation is usually cast as either a binary (\NREA) or a multiclass (\NREB) classification task.
In contrast to the binary classification framework, the current formulation of the multiclass version has an intrinsic and unknown bias term, making otherwise informative diagnostics unreliable.
We propose a multiclass framework free from the bias inherent to \NREB at optimum, leaving us in the position to run diagnostics that practitioners depend on.
It also recovers \NREA in one corner case and \NREB in the limiting case.
For fair comparison, we benchmark the behavior of all algorithms in both familiar and novel training regimes: when jointly drawn data is unlimited, when data is fixed but prior draws are unlimited, and in the commonplace fixed data and parameters setting. 
Our investigations reveal that the highest performing models are distant from the competitors (\NREA, \NREB) in hyperparameter space. We make a recommendation for hyperparameters distinct from the previous models. We suggest two bounds on the mutual information as performance metrics for simulation-based inference methods, without the need for posterior samples, and provide experimental results. This version corrects a minor implementation error in $\gamma$, improving results.
\end{abstract}

\section{Introduction}
\begin{wrapfigure}[21]{r}{0.5\textwidth}
    \centering
    \vspace{-18pt}
    \includegraphics[width=0.5\textwidth]{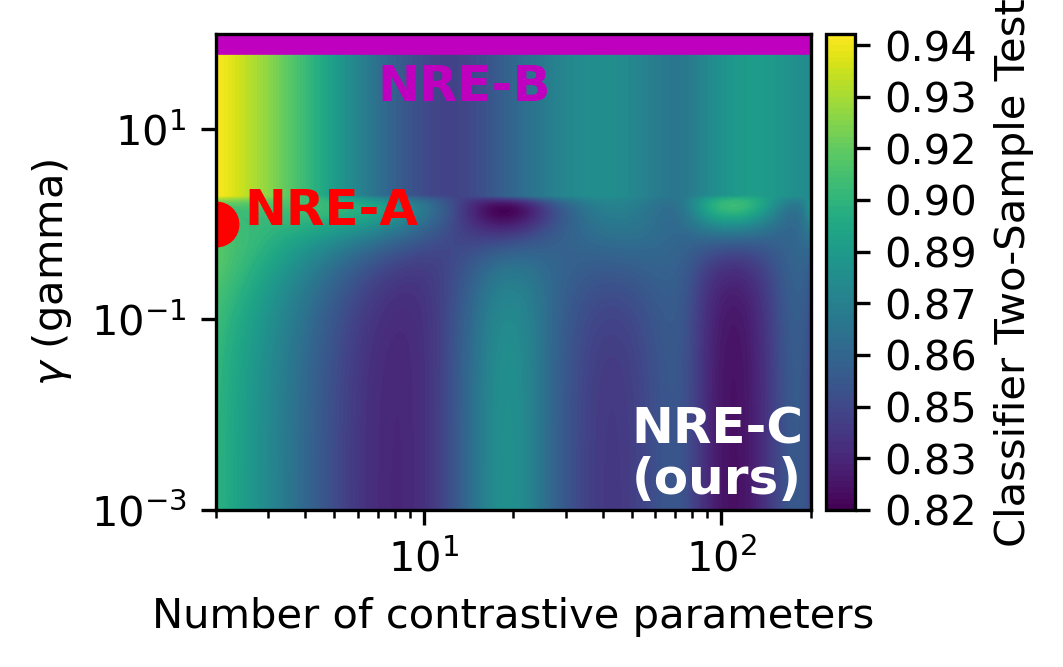}
    \caption{
        Conceptual, interpolated map from investigated hyperparameters of proposed algorithm \NREC to a measurement of posterior exactness using the Classifier Two-Sample Test. Best 0.5, worst 1.0. Red dot indicates \NREA's hyperparameters, $\gamma = 1$ and $K = 1$ \cite{Hermans2019}. Purple line implies \NREB \cite{Durkan2020} with $\gamma = \infty$ and $K \geq 1$. \NREC covers the entire plane, generalizing other methods. Best performance occurs with $K > 1$ and $\gamma \approx 1$, in contrast with the settings of existing algorithms.
    }
    \label{fig:conceptual-hyperparameters-to-c2st}
\end{wrapfigure}

We begin with a motivating example: Consider the task of inferring the mass of an exoplanet $\btheta_o$ from the light curve observations $\bx_o$ of a distant star. We design a computer program that maps hypothetical mass $\btheta$ to a simulated light curve $\bx$ using relevant physical theory. Our simulator computes $\bx$ from $\btheta$, but the inverse mapping is unspecified and likely intractable. \emph{Simulation-based inference} (\SBI) puts this problem in a probabilistic context \cite{sisson2018handbook, Cranmer2020}. Although we cannot analytically evaluate it, we assume that the simulator is sampling from the conditional probability distribution $p(\bx \mid \btheta)$. After specifying a prior $p(\btheta)$, the inverse amounts to estimating the posterior $p(\btheta \mid \bx_o)$. This problem setting occurs across scientific domains \cite{cole2021fast, alsing2018massive, brehmer2018constraining, hermans2020towards, lensing} where $\btheta$ generally represents input parameters of the simulator and $\bx$ the simulated output observation. Our design goal is to produce a surrogate model $\phat(\btheta \mid \bx)$ approximating the posterior for any data $\bx$ while limiting excessive simulation.

\begin{wrapfigure}[32]{R}{0.5\textwidth}
    \centering
    \vspace{-15pt}
    \includegraphics[width=0.5\textwidth]{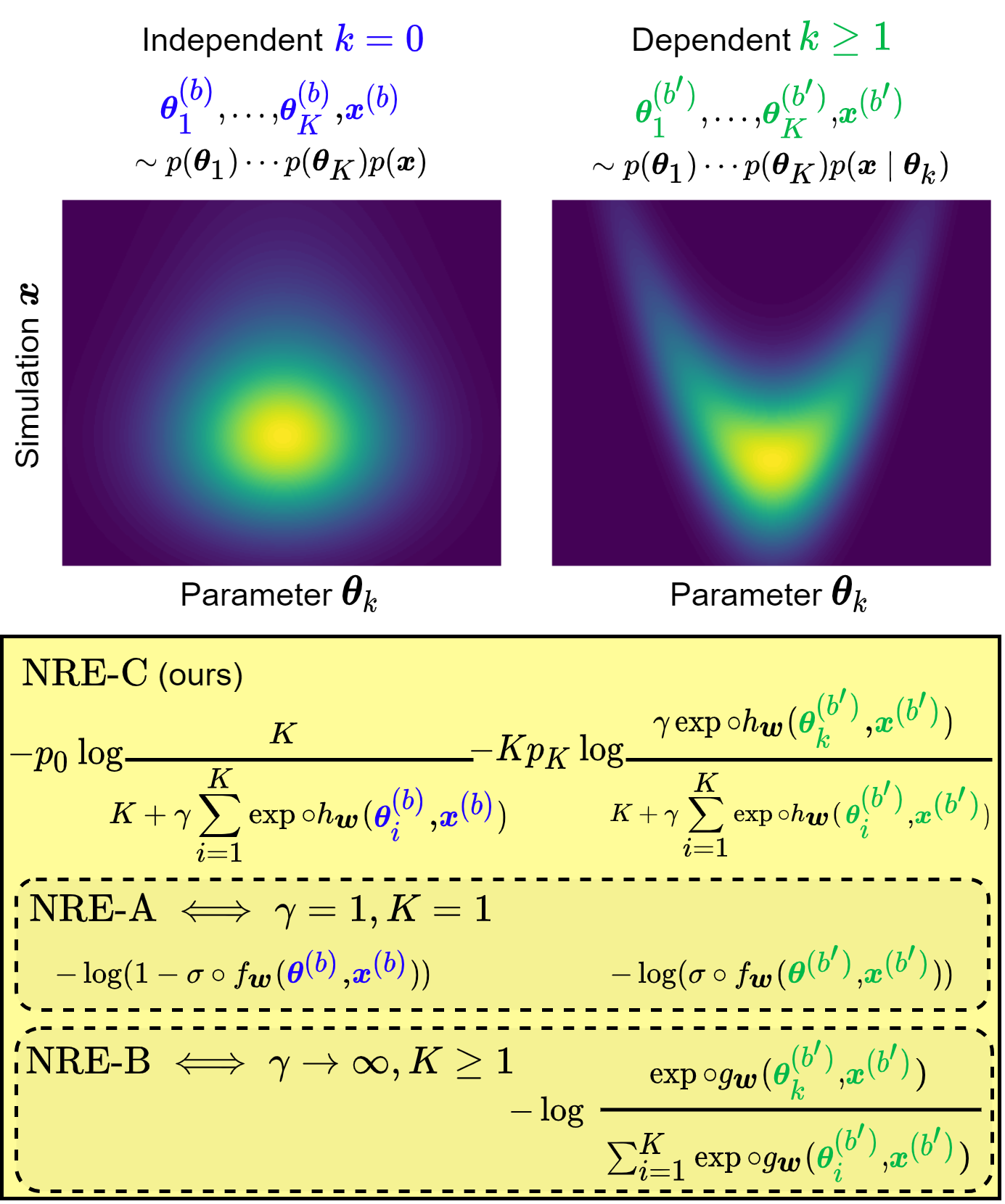}
    \caption{
        Schematic depicts how the loss is computed in \NRE algorithms. ($\btheta$, $\bx$) pairs are sampled from distributions at the top of the figure, entering the loss functions as depicted. \NREC controls the number of contrastive classes with $K$ and the weight of independent and dependent terms with $p_0$ and $p_K$. \NREC generalizes other algorithms. Hyperparameters recovering \NREA and \NREB are listed next to the name within the dashed areas. Notation details are defined in Section~\ref{sec:nrec}.
    }
    \label{fig:loss-diagram}
\end{wrapfigure}

Density estimation \cite{bishop, papamakarios2017masked, papamakarios2019normalizing} can fit the likelihood \cite{drovandi2018approximating, papamakarios2019sequential, alsing2019fast, lueckmann2019likelihood} or posterior \cite{blum2010non, papamakarios2016fast, lueckmann2017flexible, greenberg2019automatic} directly; however, an appealing alternative for practitioners is estimating a \emph{ratio} between distributions \cite{izbicki2014high, Cranmer2015, thomas2016likelihood, Hermans2019, Durkan2020}. Specifically, the likelihood-to-evidence ratio $\frac{p(\btheta \mid \bx)}{p(\btheta)} = \frac{p(\bx \mid \btheta)}{p(\bx)} = \frac{p(\btheta, \bx)}{p(\btheta) p(\bx)}$.
Unlike the other methods, ratio estimation enables easy aggregation of independent and identically drawn data $\bx$. Ratio and posterior estimation can compute bounds on the mutual information and an importance sampling diagnostic.

Estimating $\frac{p(\bx \mid \btheta)}{p(\bx)}$ can be formulated as a binary classification task \cite{Hermans2019}, where the classifier $\sigma \circ f_{\bw}(\btheta,\bx)$ distinguishes between pairs $(\btheta, \bx)$ sampled either from the joint distribution $p(\btheta, \bx)$ or the product of its marginals $p(\btheta) p(\bx)$. 
We call it \NREA. The optimal classifier has
\begin{align}
     f_{\bw}(\btheta,\bx) \approx \log \frac{p(\btheta \mid \bx)}{p(\btheta)}.
\end{align}
Here, $\sigma$ represents the sigmoid function, $\circ$ implies function composition, and $f_{\bw}$ is a neural network with weights $\bw$. As a part of an effort to unify different \SBI methods and to improve simulation-efficiency, \citet{Durkan2020} reformulated the classification task to identify which of $K$ possible $\btheta_k$ was responsible for simulating $\bx$. We refer to it as \NREB.
At optimum
\begin{align}
    g_{\bw}(\btheta, \bx) \approx \log \frac{p(\btheta \mid \bx)}{p(\btheta)} + c_{\bw}(\bx),
\end{align}
where an additional bias, $c_{\bw}(\bx)$, appears. $g_{\bw}$ represents another neural network. The $c_{\bw}(\bx)$ term nullifies many of the advantages ratio estimation offers. $c_{\bw}(\bx)$ can be arbitrarily pathological in $\bx$, meaning that the normalizing constant can take on extreme values. This limits the applicability of verification tools like the importance sampling-based diagnostic in Section~\ref{sec:diagnostic}. 

The $c_{\bw}(\bx)$ term also arises in contrastive learning \cite{gutmann2012noise, van2018representation} with \citet{ma2018noise} attempting to estimate it in order to reduce its impact. We will propose a method that discourages this bias instead. Further discussion in Appendix~\ref{apndx:mutual-information}.

There is a distinction in deep learning-based \SBI between \emph{amortized} and \emph{sequential} algorithms which produce surrogate models that estimate any posterior $p(\btheta \mid \bx)$ or a specific posterior $p(\btheta \mid \bx_o)$ respectively. Amortized algorithms sample parameters from the prior, while sequential algorithms use an alternative proposal distribution--increasing efficiency at the expense of flexibility. Amortization is usually necessary to compute diagnostics that do not require samples from $p(\btheta \mid \bx_o)$ and amortized estimators are empirically more reliable \cite{hermans2021averting}. Our study therefore focuses on amortized algorithms.

\paragraph{Contribution}
We design a more general formulation of likelihood-to-evidence ratio estimation as a multiclass problem in which the bias inherent to \NREB is discouraged by the loss function and it does not appear at optimum. Figure~\ref{fig:conceptual-hyperparameters-to-c2st} diagrams the interpolated performance as a function of hyperparameters. It shows which settings recover \NREA and \NREB, also indicating that highest performance occurs with settings distant from these. Figure~\ref{fig:loss-diagram} shows the relationship of the loss functions. We call our framework \NREC \footnote{The code for our project can be found at \href{https://github.com/bkmi/cnre}{https://github.com/bkmi/cnre} under the Apache License 2.0.} and expound the details in Section~\ref{sec:methods}.

An existing importance sampling diagnostic \cite{Hermans2019} tests whether a classifier can distinguish $p(\bx \mid \btheta)$ samples from from samples from $p(\bx)$ weighted by the estimated ratio. We demonstrate that, when estimating accurate posteriors, our proposed \NREC passes this diagnostic while \NREB does not.

Taking inspiration from mutual information estimation \cite{poole2019variational}, we propose applying a variational bound on the mutual information between $\btheta$ and $\bx$ in a novel way--as an informative metric measuring a lower bound on the Kullback-Leibler divergence between surrogate posterior estimate $p_{\bw}(\btheta \mid \bx)$ and $p(\btheta \mid \bx)$, averaged over $p(\bx)$. Unlike with two-sample testing methods commonly used in machine learning literature \cite{sbibm}, our metric samples only from $p(\btheta, \bx)$, which is always available in \SBI, and does not require samples from the intractable $p(\btheta \mid \bx)$. Our metric is meaningful to scientists working on problems with intractable posteriors. The technique requires estimating the partition function, which can be expensive. We find the metric to be well correlated with results from two-sample tests.

We evaluate \NREB and \NREC in a fair comparison in several training regimes in Section~\ref{sec:experiments}. We perform a hyperparameter search on three simulators with tractable likelihood by benchmarking the behavior when (a) jointly drawn pairs $(\btheta, \bx)$ are unlimited or when jointly drawn pairs $(\btheta, \bx)$ are fixed but we (b) can draw from the prior $p(\btheta)$ without limit or (c) are restricted to the initial pairs. We also perform the \SBI benchmark of \citet{sbibm} with our recommended hyperparameters.

\section{Methods}
\label{sec:methods}
The ratio between probability distributions can be estimated using the ``likelihood ratio trick'' by training a classifier to distinguish samples \cite{hastie2009elements, sugiyama2012density, goodfellow2014generative, Cranmer2015, thomas2016likelihood, Mohamed2016, Hermans2019}. We first summarize the loss functions of \NREA and \NREB which approximate the \emph{intractable} likelihood-to-evidence ratio $r(\bx \mid \btheta) \coloneqq \frac{p(\bx \mid \btheta)}{p(\bx)}$. We then elaborate on our proposed generalization, \NREC. Finally, we explain how to recover \NREA and \NREB within our framework and comment on the normalization properties. 

\paragraph{\NREA} \citet{Hermans2019} train a binary classifier to distinguish $(\btheta, \bx)$ pairs drawn dependently $p(\btheta, \bx)$ from those drawn independently $p(\btheta) p(\bx)$. This classifier is parameterized by a neural network $f_{\bw}$ which approximates $\log r(\bx \mid \btheta)$. We seek optimal network weights

\begin{equation}
    \bw^{*} \in \argmin_{\bw} - \frac{1}{2 B}
    \left[ \sum_{b=1}^{B} \log \left(
        1 - \sigma \circ f_{\bw}(\btheta^{(b)}, \bx^{(b)})
    \right) 
    + \sum_{b'=1}^{B}\log \left(
        \sigma \circ f_{\bw}(\btheta^{(b')}, \bx^{(b')})
    \right) \right]
\end{equation}

$\btheta^{(b)}, \bx^{(b)} \sim p(\btheta) p(\bx)$ and $\btheta^{(b')}, \bx^{(b')} \sim p(\btheta, \bx)$ over $B$ samples. 
\NREA's ratio estimate converges to $f_{\bw^*} = \log \frac{p(\bx \mid \btheta)}{p(\bx)}$ given unlimited model flexibility and data. Details can be found in Appendix~\ref{apndx:other-sbi}.
\looseness=-1

\paragraph{\NREB}
\citet{Durkan2020} train a classifier that selects from among $K$ parameters $(\btheta_1, \ldots, \btheta_K)$ which could have generated $\bx$, in contrast with \NREA's binary possibilities. One of these parameters $\btheta_k$ is \emph{always} drawn jointly with $\bx$. The classifier is parameterized by a neural network $g_{\bw}$ which approximates $\log r(\bx \mid \btheta)$. Training is done over $B$ samples by finding

\begin{equation}
\label{eqn:nreb-loss}
    \bw^* \in \argmin_{\bw}
    	\left[ - \frac{1}{B} \sum_{b'=1}^{B} \log \frac{ \exp \circ g_{\bw}(\btheta_{k}^{(b')}, \bx^{(b')}) }{ \sum_{i=1}^{K} \exp \circ g_{\bw}(\btheta_i^{(b')}, \bx^{(b')}) } \right]
\end{equation}

where $\btheta_{1}^{(b')}, \ldots, \btheta_{K}^{(b')} \sim p(\btheta)$ and $\bx^{(b')} \sim p(\bx \mid \btheta_k^{(b')})$. Given unlimited model flexibility and data \NREB's ratio estimate converges to ${g_{\bw^*}(\btheta, \bx) = \log \frac{ p(\btheta \mid \bx) }{p(\btheta)} + c_{\bw^*}(\bx)}$. Details are in Appendix~\ref{apndx:other-sbi}.
\looseness=-1

\subsection{Contrastive Neural Ratio Estimation}
\label{sec:nrec}
Our proposed algorithm \NREC trains a classifier to identify which $\btheta$ among $K$ candidates is responsible for generating a given $\bx$, inspired by \NREB. We added another option that indicates $\bx$ was drawn independently, inspired by \NREA. The introduction of the additional class yields a ratio without the specific $c_{\bw}(\bx)$ bias at optimum. Define $\bTheta \coloneqq (\btheta_1, ..., \btheta_K)$ and conditional probability

\begin{equation}
    p_{\NREC}(\bTheta, \bx \mid y = k) \coloneqq
    \begin{cases}
        p(\btheta_1) \cdots p(\btheta_K) p(\bx) & k=0 \\
    	p(\btheta_1) \cdots p(\btheta_K) p(\bx \mid \btheta_k) & k = 1, \ldots, K
    \end{cases}.
\end{equation}

We set marginal probabilities $p(y = k) \coloneqq p_{\joint}$ for all $k \geq 1$ and ${p(y = 0) \coloneqq p_{\marginal}}$, yielding the relationship $p_{\marginal} = 1 - K p_{\joint}$. Let the odds of any pair being drawn dependently to completely independently be $\gamma \coloneqq \frac{K p_{\joint}}{p_{\marginal}}$. We now use Bayes' formula to compute the conditional probability

\begin{equation}
\begin{aligned}
\label{eqn:cnre-posterior}
    p(y=k \mid \bTheta, \bx)
    &= \frac{p(y=k) \, p(\bTheta, \bx \mid y=k)/p(\bTheta, \bx \mid y=0)}{\sum_{i=0}^{K} p(y=i) \,  p(\bTheta, \bx \mid y=i)/p(\bTheta, \bx \mid y=0)} \\
    &= \frac{p(y=k) \, p(\bTheta, \bx \mid y=k)/p(\bTheta, \bx \mid y=0)}{p(y=0) + \sum_{i=1}^{K} p(y=i) \, p(\bTheta, \bx \mid y=i)/p(\bTheta, \bx \mid y=0)} \\
    &= \begin{cases}
    	\frac{K}{K + \gamma \sum_{i=1}^{K} r(\bx \mid \btheta_i)} & k=0 \\
    	\frac{\gamma \, r(\bx \mid \btheta_k)}{K + \gamma \sum_{i=1}^{K} r(\bx \mid \btheta_i)} & k=1, \ldots, K
    \end{cases}.
\end{aligned}
\end{equation}

We dropped the \NREC subscript and substituted in $\gamma$ to replace the $p(y)$ class probabilities. We train a classifier, parameterized by neural network $h_{\bw}(\btheta, \bx)$ with weights $\bw$, to approximate \eqref{eqn:cnre-posterior} by

\begin{equation}
\label{eqn:classifier}
    q_{\bw}(y = k \mid \bTheta, \bx) =
    \begin{cases}
    	\frac{K}{K + \gamma \sum_{i=1}^{K} \exp \circ h_{\bw}(\btheta_i, \bx)} & k = 0 \\
    	\frac{\gamma \, \exp \circ h_{\bw}(\btheta_k,\bx))}{K + \gamma \sum_{i=1}^{K} \exp \circ h_{\bw}(\btheta_i,\bx)} & k = 1, \ldots, K.
    \end{cases}.
\end{equation}

We note that \eqref{eqn:classifier} still satisfies $\sum_{k=0}^K q_{\bw}(y = k \mid \bTheta, \bx) =1$, no matter the parameterization.

\paragraph{Optimization}
We design a loss function that encourages $h_{\bw}(\btheta, \bx) = \log \frac{p(\bx \mid \btheta)}{p(\bx)}$ at convergence, and holds at optimum with unlimited flexibility and data. We introduce the cross entropy loss
\begin{equation}
\begin{aligned}
    \ell(\bw) 
    &\coloneqq \mathbb{E}_{p(y, \bTheta, \bx)} \left[ -\log q_{\bw}(y \mid \bTheta, \bx)\right] \\
    &= - p_{\marginal} \mathbb{E}_{p(\bTheta, \bx \mid y=0)} \left[
        \log q_{\bw}(y =0 \mid \bTheta, \bx) 
    \right] 
    - p_{\joint} \sum_{k=1}^{K} \mathbb{E}_{p(\bTheta, \bx \mid y=k)} \left[ 
        \log q_{\bw}(y = k \mid \bTheta, \bx) 
    \right] \\
    &= - p_{\marginal} \mathbb{E}_{p(\bTheta, \bx \mid y=0)} \left[
    \log q_{\bw}(y =0 \mid \bTheta, \bx) 
    \right] 
    - K p_{\joint} \mathbb{E}_{p(\bTheta, \bx \mid y=K)} \left[ 
    \log q_{\bw}(y = K \mid \bTheta, \bx) 
    \right]
\end{aligned}
\end{equation}
and minimize it towards $\bw^* \in \argmin_{\bw} \ell(\bw)$. 
We point out that the final term is symmetric up to permutation of $\bTheta$, enabling the replacement of the sum by multiplication with $K$. When $\gamma$ and $K$ are known, $p_{\marginal} = \frac{1}{1 + \gamma}$ and $p_{\joint} = \frac{1}{K} \frac{\gamma}{1 + \gamma}$ under our constraints. Without loss of generality, we let $\btheta_1, \ldots, \btheta_{K} \sim p(\btheta)$ and  $\bx \sim p(\bx \mid \btheta_{K})$. An empirical estimate of the loss on $B$ samples is therefore
\begin{equation}
\label{eqn:nrec-empirical-loss}
\begin{aligned}
    \hat{\ell}_{\gamma, K}(\bw) 
    &\coloneqq - \frac{1}{B} 
    \Bigg[ \frac{1}{1 + \gamma} \sum_{b=1}^{B} \log q_{\bw} \left(y = 0 \mid \bTheta^{(b)}, \bx^{(b)} \right) \\
    &\phantom{\approx - \frac{1}{B} \Bigg[ } + \frac{\gamma}{1 + \gamma} \sum_{b'=1}^{B} \log q_{\bw} \left(y = K \mid \bTheta^{(b')}, \bx^{(b')} \right) \Bigg].
\end{aligned}
\end{equation}

In the first term, the classifier sees a completely independently drawn sample of $\bx$ and $\bTheta$ while $\btheta_K$ is drawn jointly with $\bx$ in the second term. In both terms, the classifier considers $K$ choices. In practice, we bootstrap both $\btheta_1^{(b)}, \ldots, \btheta_{K}^{(b)}$ and $\btheta_1^{(b')}, \ldots, \btheta_{K-1}^{(b')}$ from the same mini-batch and compare them to the same $\bx$, similarly to \NREA and \NREB. Proof of the above is in Appendix~\ref{apndx:proof}.

\paragraph{Recovering \NREA and \NREB}
\NREC is general because specific hyperparameter settings recover \NREA and \NREB. To recover \NREA one should set $\gamma = 1$ and $K=1$ in \eqref{eqn:nrec-empirical-loss} yielding 

\begin{equation}
\begin{aligned}
    \hat{\ell}_{1, 1}(\bw) 
    &= - \frac{1}{2 B} 
    \Bigg[
    \sum_{b=1}^{B} \log \frac{1}{1 + \exp \circ h_{\bw}(\btheta^{(b)}, \bx^{(b)})}
    + \sum_{b'=1}^{B}  \log \frac{\exp \circ h_{\bw}(\btheta^{(b')}, \bx^{(b')})}{1 + \exp \circ h_{\bw}(\btheta^{(b')}, \bx^{(b')})}
    \Bigg] \\
    &= - \frac{1}{2B} \left[- \sum_{b=1}^{B} \log \left(
        1 - \sigma \circ h_{\bw}(\btheta^{(b)}, \bx^{(b)})
    \right) 
    + \sum_{b'=1}^{B}\log \left(
        \sigma \circ h_{\bw}(\btheta^{(b')}, \bx^{(b')})
    \right) \right]
\end{aligned}
\end{equation}
where we dropped the lower index. Recovering \NREB requires taking the limit $\gamma \to \infty$ in the loss function. In that case, the first term goes to zero, and second term converges to the softmax function.
\looseness=-1

\begin{equation}
\begin{aligned}
    \hat\ell_{\infty, K}(\bw)
    &= \lim_{\gamma \to \infty} \hat\ell_{\gamma, K}(\bw)
    &= - \frac{1}{B} 
    \Bigg[ \sum_{b'=1}^{B} \log 
    \frac{\exp \circ h_{\bw}(\btheta_k,\bx))}{\sum_{i=1}^{K} \exp \circ h_{\bw}(\btheta_i,\bx)}
    \Bigg]
\end{aligned}
\end{equation}

is determined by substitution into \eqref{eqn:nrec-empirical-loss}. Both equations are obviously the same as their counterparts.

\paragraph{Estimating a normalized posterior} 
In the limit of infinite data and infinite neural network capacity (width, depth) the optimal classifier trained using \NREC (with $\gamma \in \mathbb{R}^{+}$) satisfies the equality:
\begin{align}
    h_{\bw^*}(\btheta,\bx) & = \log \frac{p(\btheta\mid\bx)}{p(\btheta)}.
\end{align}
In particular, we have that the following normalizing constant is trivial:
\begin{align}
  Z(\bx) & := \int \exp\left(h_{\bw^*}(\btheta,\bx)\right) p(\btheta)\,d\btheta = \int p(\btheta\mid\bx)\,d\btheta =  1.
\end{align}
This is a result of Lemma~\ref{lem:optimal=log-ratio} in Appendix~\ref{apndx:proof}. However, practitioners never operate in this setting, rather they use finite sample sizes and neural networks with limited capacity that are optimized locally. The non-optimal function $\exp(h_{\bw}(\btheta,\bx))$ does not have a direct interpretation as a ratio of probability distributions, rather as the function to weigh the prior $p(\btheta)$ to approximate the unnormalized posterior. In other words, we find the following approximation for the posterior $p(\btheta\mid\bx)$:
\begin{align}
   \label{eqn:normalizing-constant}
   p_{\bw}(\btheta\mid\bx) &:= \frac{\exp(h_{\bw}(\btheta,\bx))}{Z_{\bw}(\bx)} p(\btheta), &
  Z_{\bw}(\bx) & := \int \exp\left(h_{\bw}(\btheta,\bx)\right) p(\btheta)\, d\btheta,
\end{align}
where in general the normalizing constant is not trivial, i.e.\ $Z_{\bw}(\bx) \neq 1$. As stated above, the \NREC (and \NREA) objective encourages $Z_{\bw}(\bx)$ to converge to $1$. This is in sharp contrast to  \NREB, where even at optimum with an unrestricted function class a non-trivial $\bx$-dependent bias term can appear. 

There is no restriction on how pathological the \NREB bias $c_{\bw}(\bx)$ can be. Consider a minimizer of \eqref{eqn:nreb-loss}, the \NREB loss function, $h_{\bw^{\ast}} + c_{\bw^{\ast}}(\bx)$. Adding any function $d(\bx)$ cancels out in the fraction and is also a minimizer of \eqref{eqn:nreb-loss}. This freedom complicates any numerical computation of the normalizing constant and renders the importance sampling diagnostic from Section~\ref{sec:diagnostic} generally inapplicable. We report Monte Carlo estimates of $Z_{\bw}(\bx)$ on a test problem across hyperparameters in Figure~\ref{fig:partition-function}.

\subsection{Measuring performance \& ratio estimator diagnostics}
\label{sec:diagnostic}
\SBI is difficult to verify because, for many use cases, the practitioner cannot compare surrogate $p_{\bw}(\btheta \mid \bx)$ to the intractable ground truth $p(\btheta \mid \bx)$. Incongruous with the practical use case for \SBI, much of the literature has focused on measuring the similarity between surrogate and posterior using two-samples tests on tractable problems. For comparison with literature, we first reference a two-sample exactness metric which requires a tractable posterior. We then discuss diagnostics which do not require samples from $p(\btheta \mid \bx)$, commenting on the relevance for each \NRE algorithm with empirical results. Further, we find that a known variational bound to the mutual information is tractable to estimate within \SBI, that it bounds the average Kullback-Leibler divergence between surrogate and posterior, and propose to use it for model comparison on intractable inference tasks.

\paragraph{Comparing to a tractable posterior with estimates of exactness}
Assessments of approximate posterior quality are available when samples can be drawn from both the posterior $\btheta \sim p(\btheta \mid \bx)$ and the approximation $\btheta \sim q(\btheta \mid \bx)$. In the deep learning-based \SBI literature, exactness is measured as a function of computational cost, usually simulator calls. We investigate this with \NREC in Section~\ref{sec:benchmark}.

Based on the recommendations of \citet{sbibm} our experimental results are measured using the Classifier Two-Sample Test (C2ST) \cite{friedman2003multivariate, lehmann2005testing, lopez2017revisiting}. A classifier is trained to distinguish samples from either the surrogate or the ground truth posterior. An average classification probability on holdout data of 1.0 implies that samples from each distribution are easily identified; 0.5 implies either the distributions are the same or the classifier does not have the capacity to distinguish them.

\paragraph{Importance sampling diagnostic}
\begin{figure}[hbt]
\centering
    \begin{subfigure}[b]{0.3\textwidth}
        \includegraphics[width=\textwidth]{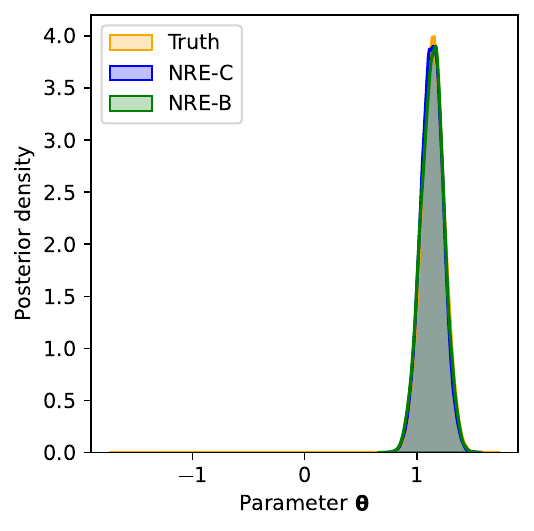}
        \caption{Posteriors}
        \end{subfigure}
    \hfill
        \begin{subfigure}[b]{0.3\textwidth}
        \includegraphics[width=\textwidth]{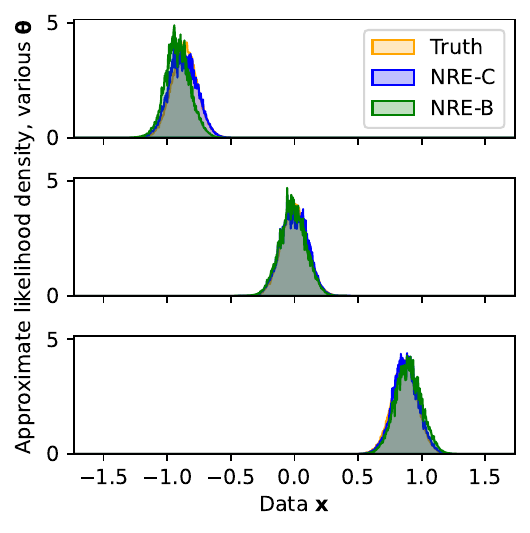}
        \caption{Likelihood and reweighted marginal generative models.}
        \end{subfigure}
    \hfill
        \begin{subfigure}[b]{0.3\textwidth}
        \includegraphics[width=\textwidth]{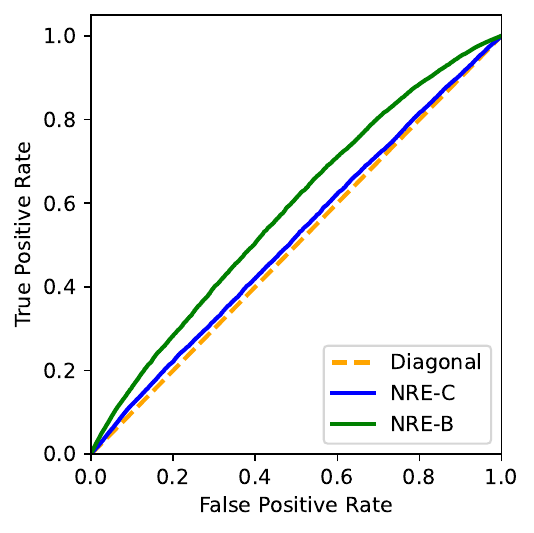}
        \caption{Receiver operating characteristic.}
    \end{subfigure}
    \caption{
        The figures visualize the importance sampling diagnostic on ratio estimators trained using \NREB and \NREC. (a) Both methods produce satisfactory posterior estimates that agree with $p(\btheta \mid \bx)$. (b) $p(\bx \mid \btheta)$ is shown along with $p(\bx)$ samples weighted by \NREA $\exp \circ f_{\bw}(\btheta, \bx)$ and \NREB $\exp \circ g_{\bw}(\btheta, \bx)$. Each plot corresponds to a different $\btheta$. Despite high posterior accuracy, the \NREB estimates are distinct from $p(\bx \mid \btheta)$. (c) Two classifier's \ROC curves, each trained to distinguish $p(\bx \mid \btheta)$ samples from $p(\bx)$ samples weighted by the corresponding \NRE's $\rhat$ estimate. The classifier failed to distinguish likelihood samples from the \NREC weighted data samples, but successfully identified \NREB weighted samples. \NREB accurately approximates the posterior, but fails the diagnostic. \NREC produces an accurate posterior surrogate and passes the diagnostic.
    }
    \label{fig:importance-sampling-diagnostic}
\end{figure}
An accurate likelihood-to-evidence weight transforms the data distribution into the likelihood by ${p(\bx \mid \btheta) = p(\bx) r(\bx \mid \btheta)}$. Since \NRE necessitates simulator access, we can test the ratio estimator by training a classifier to distinguish unweighted $p(\bx \mid \btheta)$ samples from weighted $p(\bx) \rhat(\bx \mid \btheta)$ samples, where $\rhat$ implies an estimate. Indistinguishability between samples implies either that the approximate ratio is accurate for parameter $\btheta$ or that the classifier does not have sufficient power to find predictive features. Issues with classification power can be detected by assessing the classifier's ability to distinguish $p(\bx)$ from $p(\bx \mid \btheta)$. The performance can be visualized in a receiver operating curve (\ROC) or measured by the area under the curve (\ROC \AUC). This diagnostic has been used for ratio estimators before \cite{Cranmer2015, Hermans2019} but it comes from training models under covariate shift \cite{shimodaira2000improving}. It is particularly appealing because it does not require samples from $p(\btheta \mid \bx)$.
\looseness=-1

\citet{Durkan2020} do not mention this diagnostic in their paper, but due to its intrinsic bias \NREB does not fulfill the identity necessary for this diagnostic to hold at optimum. The unknown factor that depends on $\bx$ implies ${p(\bx \mid \btheta) \neq p(\bx) \exp\circ g_{\bw}(\bx \mid \btheta)}$. We provide empirical evidence of this issue in Figure~\ref{fig:importance-sampling-diagnostic}. Although \NREB accurately approximates the true posterior, it demonstrably fails the diagnostic. Given the limited options for verification of \SBI results, this presents a major problem by significantly limiting the trustworthiness of \NREB on any problem with an intractable posterior. In Appendix~\ref{apndx:proof}, we show that the unrestricted \NREB-specific $c_{\bw}(\bx)$ bias means approximating $p(\bx \mid \btheta)$ with normalized importance weights will not solve the issue.

\paragraph{Mutual information bound}
Selecting the surrogate model most-similar to the target posterior remains intractable without access to $p(\btheta \mid \bx_o)$. Nevertheless, practitioners must decide which surrogate should approximate the posterior across training and hyperparameter search. Unfortunately, the validation losses between different versions of \NRE and different $K$ and $\gamma$ settings are not comparable. A good heuristic is to choose the model which minimizes the Kullback-Leibler divergence \emph{on average} over possible data $p(\bx)$. 
In Appendix~\ref{apndx:mutual-information}, we prove the relationship between $I(\btheta; \bx)$, the \emph{mutual information} with respect to $p(\btheta, \bx)$, a pair of variational bounds on the mutual information $I_{\bw}^{(0)}(\btheta; \bx)$ and $I_{\bw}^{(1)}(\btheta; \bx)$, and the average \KLD
\begin{equation}
\label{eqn:average-kld-is-mi}
    \mathbb{E}_{p(\bx)} \left[ \KLD(p(\btheta \mid \bx) \Mid p_{\bw}(\btheta \mid \bx)) \right] 
    = I(\btheta; \bx) - I_{\bw}^{(0)}(\btheta; \bx),
\end{equation}
\begin{equation}
\label{eqn:define-mi-0-in-paper}
    I_{\bw}^{(0)}(\btheta; \bx)
    \coloneqq \mathbb{E}_{p(\btheta, \bx)} \left[ \log \rhat(\bx \mid \btheta) \right] - \mathbb{E}_{p(\bx)} \left[ \log \mathbb{E}_{p(\btheta)} [\rhat(\bx \mid \btheta)] \right].
\end{equation}
\begin{equation}
\label{eqn:define-mi-1-in-paper}
    I_{\bw}^{(1)}(\btheta; \bx) 
    \coloneqq 
	\mathbb{E}_{p(\btheta, \bx)} \left[ \log \rhat(\bx \mid \btheta) \right] - \mathbb{E}_{p(\bx)p(\btheta)} \left[ \rhat(\bx \mid \btheta) -1 \right].
\end{equation}
The non-negativity of all terms in \eqref{eqn:average-kld-is-mi} implies $I(\btheta; \bx) \geq I_{\bw}^{(0)}(\btheta; \bx)$; that means the model which minimizes $-I_{\bw}^{(0)}(\btheta; \bx)$ best satisfies our heuristic. Furthermore, in Appendix~\ref{apndx:mutual-information} we show $I_{\bw}^{(0)}(\btheta; \bx) \geq I_{\bw}^{(1)}(\btheta; \bx)$.
We propose to approximate $-I_{\bw}^{(0)}(\btheta; \bx)$ or $-I_{\bw}^{(1)}(\btheta; \bx)$ with Monte Carlo using held-out data as a metric for model selection during training and across hyperparameters. The expectation values average over $p(\btheta, \bx)$, $p(\btheta)$, and $p(\bx)$. We can sample from all of these distributions in the \SBI context. Since the second term in \eqref{eqn:define-mi-0-in-paper} computes the average log partition function, our metric can compare \NREB-based surrogates to \NREC-based ones. 
However, the metric comes with the normal challenges of estimating the log partition function, which can be very expensive. Naively estimating the log partition function on samples is biased do to the log in the second term on the R.H.S. of \eqref{eqn:define-mi-0-in-paper}. Although $-I_{\bw}^{(1)}(\btheta; \bx)$ is unbiased, the presence of the ratio in the integrand can make the last term in the R.H.S. of \eqref{eqn:define-mi-1-in-paper} high variance. We treat a generally tractable bound in Appendix~\ref{apndx:mutual-information}. We go into \emph{much} more depth there and discuss of the relevance to Neural Posterior Estimation \cite{papamakarios2016fast, lueckmann2017flexible, greenberg2019automatic, Durkan2020}. While the application to \SBI is novel, bounds on the mutual information have been broadly investigated for contrastive representation learning and mutual information estimation \cite{gutmann2010noise, gutmann2012noise, gutmann2022statistical, belghazi2018mine, van2018representation, poole2019variational}.

\paragraph{Empirical expected coverage probability}
For a candidate distribution to qualify as the posterior, integrating over data must return the prior. A measurement that follows from calibration to the prior is called expected coverage probability. Expected coverage probability can be estimated with samples from $p(\btheta, \bx)$ and any amortized \SBI method. Although important, ability to compute this metric does not distinguish \NREC. We refer the interested reader to \citet{hermans2021averting}. We note that popular sequential techniques generally render this diagnostic inapplicable, with exceptions \cite{miller2020simulation, miller2021truncated, cole2021fast}.

\begin{wrapfigure}[39]{R}{0.5\textwidth}
    \centering
    \vspace{-13pt}
    \includegraphics[width=0.5\textwidth]{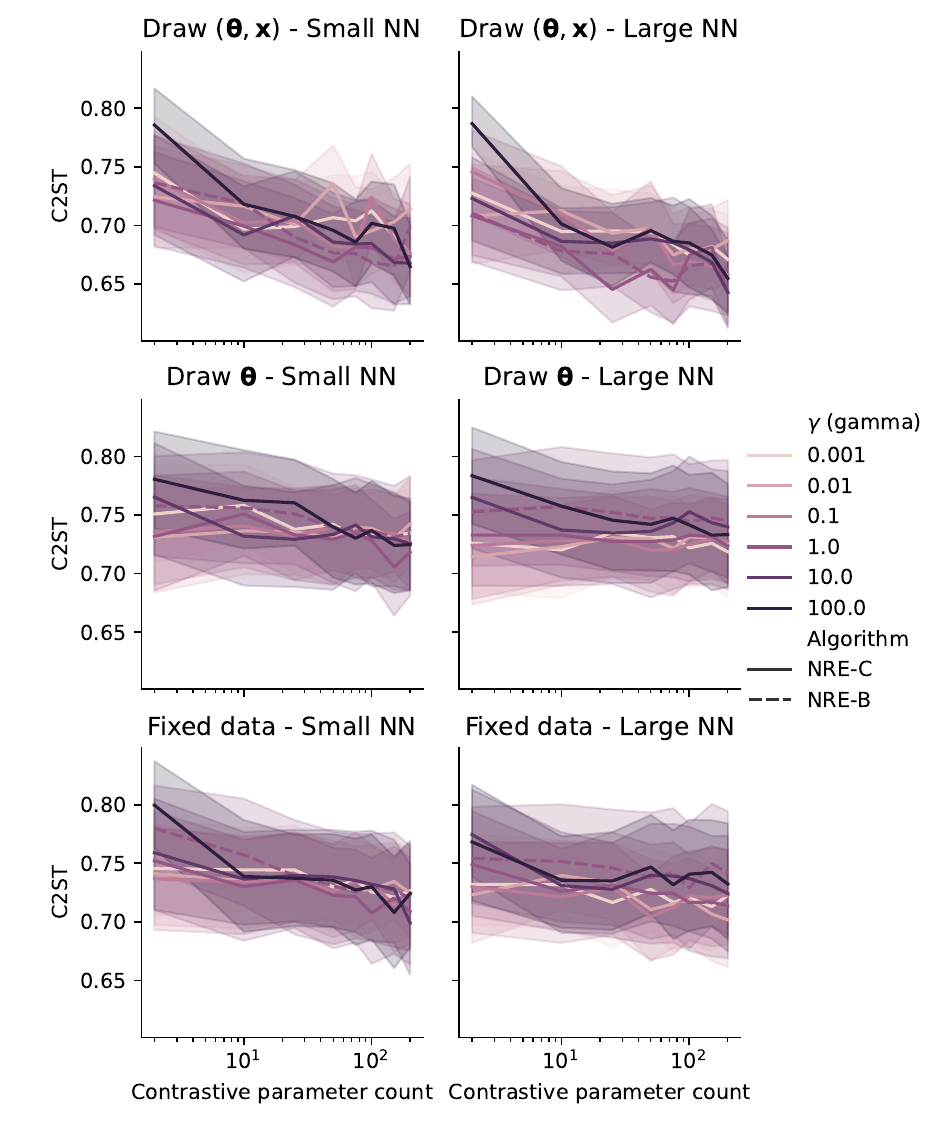}
    \caption{%
    Exactness of \NRE posterior surrogates is computed for various contrastive parameter counts, $\gamma$ values, and architectures on an \emph{average} of three tasks from the \SBI benchmark \cite{sbibm}. C2ST assigns 1.0 to inaccurate and 0.5 to accurate posterior approximation.
    (top) $p(\btheta, \bx)$ was sampled at every mini-batch during training. The accuracy strongly depends on $K$.
    (mid) A fixed number of dependent $(\btheta, \bx)$ pairs were drawn, but $p(\btheta)$ was sampled at every mini-batch during training. In this regime, $K$ has a smaller effect.
    (bot) The training data is completely fixed. Contrastive parameters are drawn in a bootstrap from the mini-batch. On the problems with fixed simulation data $\bx$, higher $K$ improves accuracy and small $\gamma$ with larger architectures slightly improves performance. The effects of the architecture are more clearly seen on difficult problems like SLCP in Appendix~\ref{apndx:experimental-details}.
    }
    \label{fig:c2st-average}
\end{wrapfigure}

\section{Experiments}
\label{sec:experiments}

We perform experiments in three settings to 
measure the exactness of surrogate models under various hyperparameters and training regimes. Section~\ref{sec:unlimited-joint} aims to identify whether data or architecture is the bottleneck in accuracy by drawing from the joint with every mini-batch. The next experiments aim to determine how to optimally extract inference information given a limited amount of simulation data. Section~\ref{sec:unlimited-prior} leverages a cheap prior by drawing new contrastive parameters with every mini-batch. Finally, Section~\ref{sec:benchmark} applies the commonplace training regime for deep learning-based \SBI literature of fixed data and bootstrapped contrastive parameters. In this setting, we also benchmark \NREC on ten inference tasks from \citet{sbibm} using our recommended hyperparameters.

On all hyperparameter searches we consider three simulators from the simulation-based inference benchmark, namely SLCP, Two Moons, and Gaussian Mixture \cite{sbibm}. SLCP is a difficult inference task which has a simple likelihood and complex posterior \cite{greenberg2019automatic, papamakarios2019sequential}. Parameters are five dimensional and the data are four samples from a two dimensional random variable. Two Moons is two dimensional with a crescent-shaped bimodal posterior. Finally, the Gaussian Mixture data draws from a mixture of two, two-dimensional normal distributions with extremely different covariances \cite{sisson2018handbook, Beaumont2009-gl, Toni2009-fd, simola2020adaptive}.

Our surrogate models are parameterized by one of these architectures: \emph{Small NN} is like the benchmark with 50 hidden units and two residual blocks. \emph{Large NN} has 128 hidden units and three residual blocks. We use batch normalization, unlike the benchmark. We compare their performance on a grid of $\gamma$ and $K$ values. We report post-training results using the C2ST, and mid-training validation loss for insight into convergence rate. We generally use residual networks \cite{resnet} with batch normalization \cite{ioffe2015batch} and train them using adam \cite{kingma2015adam}. We also run $\NREB$ with the same architecture for comparison. To compare with \NREB we set the number of total contrastive parameters equal.

What does fair comparison mean in our experimentation? We compare models across fixed number of gradient steps. This implies that models with more classes, i.e., greater $K$, evaluate the classifier-in-training on more pairs at a given training step than a model with fewer classes. An alternative which we do \emph{not} apply in this paper: Vary the number of gradient steps, holding the number of pair evaluations fixed, i.e., a model with higher $K$ sees the same number of pairs as a model with lower $K$ but the model with lower $K$ has been updated more times. We leave this analysis for future work.

\vspace{3em}

\subsection{Behavior with unlimited data}
\label{sec:unlimited-joint}
What is responsible for inaccuracies of the surrogate model when training has saturated? (a) the amount of training data (b) the flexibility of the model? In this section we provide new simulation and parameter data with every mini-batch and train until saturation, thereby eliminating the possibility of (a). The newly drawn mini-batch parameters $\bTheta$ are bootstrapped for the contrastive pairs. 

The setting is similar to \REJABC where simulations are drawn until the posterior has converged. The results of this study will provide a baseline to compare with limited-data results and help us understand how the deep learning architecture's limitations are affected by our introduced hyperparameters. The results are reported in the top row of Figure~\ref{fig:c2st-average}.

The trend is that increasing the number of contrastive examples helps \NREB and \NREC.

Best performance on the average C2ST was found when using more contrastive parameters and the higher values of $\gamma$. Much of this can be attributed to the improved performance on SLCP, as evidenced by the plot broken down by task in Figure~\ref{fig:c2st-specific}. When contrastive parameters are fixed to a low value, large values of $\gamma$ were generally harmful to C2ST performance. $\gamma = 1$ generally performed acceptably.

Study of the detailed validation losses in Appendix~\ref{apndx:experimental-details} reveals that high $\gamma$ is associated with higher variance validation loss during training. For Gaussian Mixture, saturation is reached before the maximum number of epochs have elapsed. At that level of training, saturation is beginning for the SLCP and Two Moons. SLCP may benefit the most from more training data, since it is the highest-dimensional and most complex task.

We argue that the performance bottleneck has to do with the number of contrastive parameters, network flexibility, or training details rather than the amount of data. (Except perhaps on SLCP.) Despite saturating validation losses with unlimited training data, the C2ST improves based on contrastive parameters and network flexibility. Intuitively a more flexible model could continue to extract information from this unlimited set of jointly drawn $(\btheta, \bx)$ pairs and more contrastive parameters implies more information gained per gradient step. This has some consequence because network flexibility may limit performance in the benchmark case. Appendix~\ref{apndx:experimental-details} contains detailed results.

\subsection{Leveraging fast priors (drawing theta)}
\label{sec:unlimited-prior}
In practice, the simulator is often slow but one can often draw from the prior at the same pace as training a neural network for one mini-batch. Our goal is to understand how our hyperparameter are affected by this setting and whether it is valuable to use this technique in practice. To our knowledge, this training regime has not been explored in the deep learning-based \SBI literature.

Initially we draw a fixed set of around 20,000 samples $(\btheta, \bx) \sim p(\btheta, \bx)$. For every mini-batch during training, we sample all necessary contrastive parameters from the prior $p(\btheta)$. For each term in $\hat{\ell}_{\gamma, K}$ we take the same batch of contrastive parameters and reshuffle them, thereby equalizing the number of samples seen by \NREB and \NREC, up to bootstrap. The averaged C2ST results from the three inference tasks are reported in the middle row of Figure~\ref{fig:c2st-average}. 

As seen in Figure~\ref{fig:c2st-average}, the resulting estimators are markedly less sensitive to the number of contrastive parameters than in Section~\ref{sec:unlimited-joint}. The effects of using Large NN or Small NN are limited, although Large NN does perform better with high $K$ and low $\gamma$, particularly on SLCP as seen in Figure~\ref{fig:c2st-specific}.

The validation loss curves in Appendix~\ref{apndx:experimental-details}, show higher variance than in the experiment from Section~\ref{sec:unlimited-joint}. Additionally, improved performance using Small NN given higher $K$, may be explained by an increased convergence rate.

We claim the results from this section imply that drawing contrastive parameters from the prior helps extract the maximum amount of information from fixed simulation data $\bx$--while increasing the number of contrastive parameters $K$ has only a small positive effect.

\subsection{Simulation-based inference benchmark}
\label{sec:benchmark}

In this section, we assume the traditional literature setting of limited simulation and prior budget. Once we've selected hyperparameters based on a grid search of a subset of the \SBI benchmark, we perform the entire benchmark with those hyperparameters.

\paragraph{Hyperparameter search}
To compare to the previous two sections, the amount of training data was fixed such that each epoch was comparable to Section~\ref{sec:unlimited-joint}, this amounts to about 20,000 samples. We first inspect the C2ST results in the last row of Figure~\ref{fig:c2st-average}. The sensitivity to $K$ appears less than in Section~\ref{sec:unlimited-joint} but more than in Section~\ref{sec:unlimited-prior}. Smaller $\gamma$ slightly improves Large NN's performance but the noise makes this result uncertain. The larger network performs better but this is primarily due to the SLCP task, see Appendix~\ref{apndx:experimental-details}.

\begin{figure}[htb]
\centering
    \begin{subfigure}[b]{0.218\textwidth}
        \includegraphics[width=\textwidth]{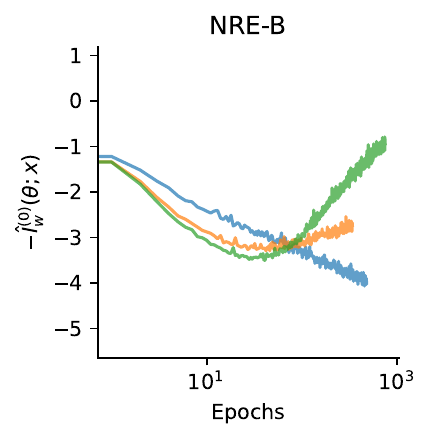}
        \includegraphics[width=\textwidth]{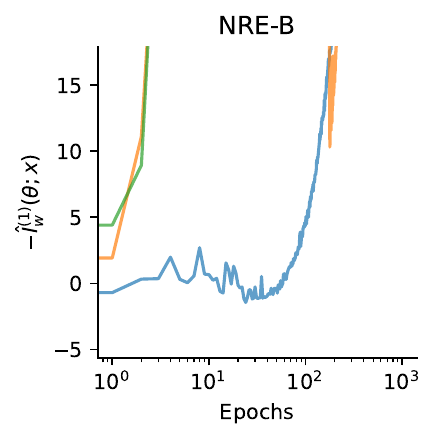}
        \caption{\NREB}
        \end{subfigure}
    \hfill
        \begin{subfigure}[b]{0.775\textwidth}
        \includegraphics[width=\textwidth]{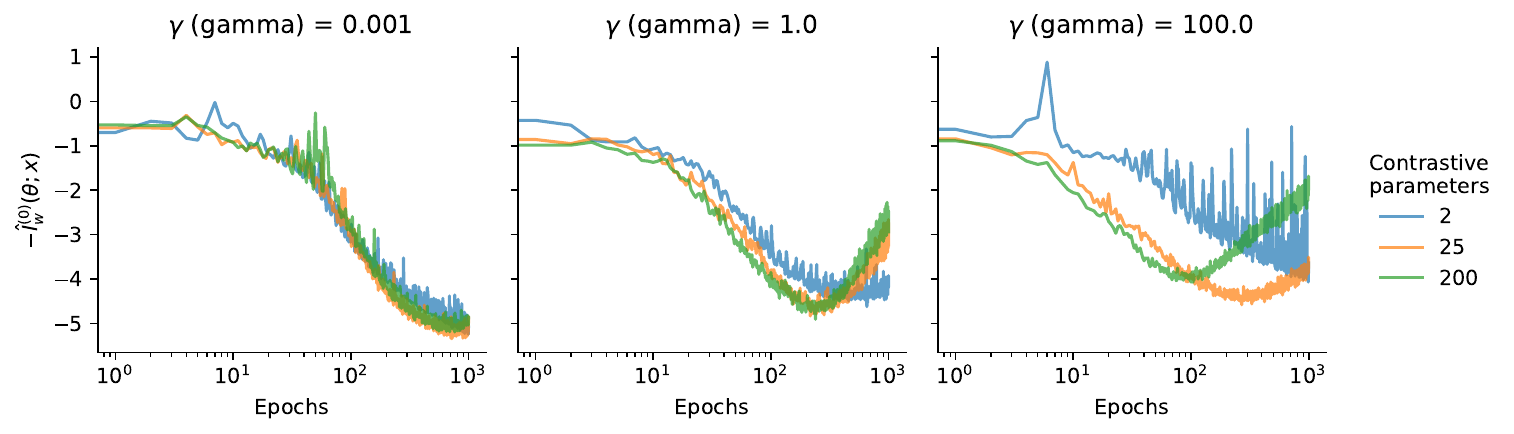}
        \includegraphics[width=\textwidth]{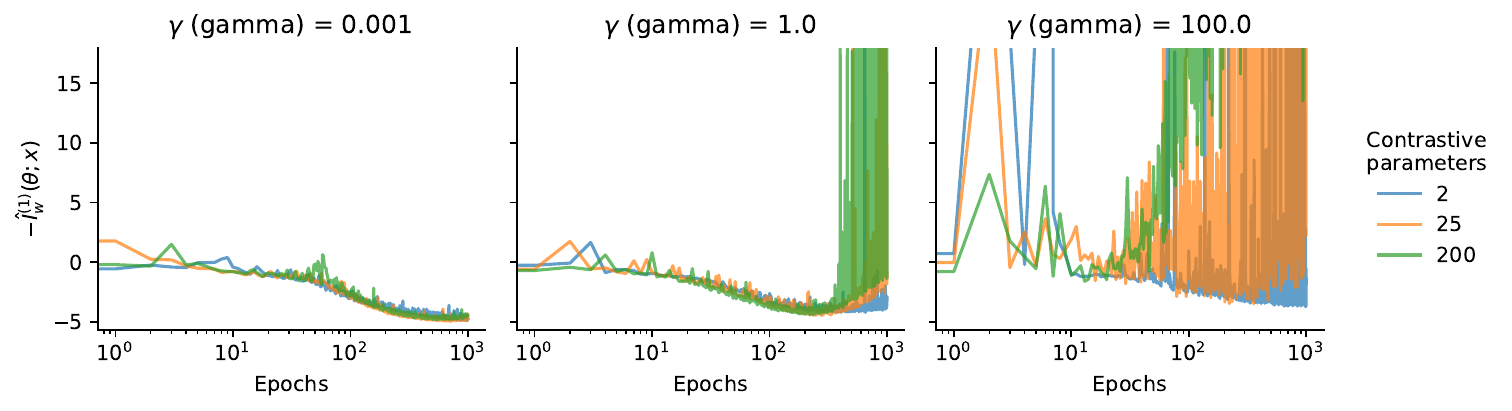}
        \caption{\NREC}
        \end{subfigure}
    \caption{
        Our proposed metrics, a pair of bounds on the mutual information $-\hat{I}_{\bw}^{(0)}(\btheta; \bx)$ (top) and $-\hat{I}_{\bw}^{(1)}(\btheta; \bx)$ (bottom), for the SLCP task estimated over the validation set versus training epochs using (a) \NREB and (b) \NREC with various values of $\gamma$ and $K$, a Large NN architecture, and fixed training data.
        Recall, the i.i.d. estimates of $-\hat{I}_{\bw}^{(0)}(\btheta; \bx)$ are biased and $-\hat{I}_{\bw}^{(1)}(\btheta; \bx)$ are high variance.
        Our conclusions will be based on $-\hat{I}_{\bw}^{(0)}(\btheta; \bx)$, which is the most readable. For both \NREB and \NREC, increasing $K$ tends to positively affect the convergence rate and optimal performance (unless $\gamma$ is very large). Increasing $\gamma$ increases convergence rate for a fixed $K$. Meanwhile, smaller $\gamma$ led to slightly better optima on this task. It's possible that the low values of $\gamma$ act as a regularizer, helping to generalize from the training data on this complex task and slowing convergence.
    }
    \label{fig:mutual-information-0-slcp-summary}
\end{figure}

We do a Monte Carlo estimate of our proposed metrics, denoted $-\hat{I}_{\bw}^{(0)}(\btheta; \bx)$ and $-\hat{I}_{\bw}^{(1)}(\btheta; \bx)$, to show performance on the SLCP task as a function of epochs in Figure~\ref{fig:mutual-information-0-slcp-summary}.
For both \NREB and \NREC, increasing $K$ tends to positively affect the convergence rate and optimal performance (unless $\gamma$ is very large). Increasing $\gamma$ increases convergence rate for a fixed $K$. Meanwhile, smaller $\gamma$ led to slightly better optima on this task, albeit with a slower convergence rate. See Figure~\ref{fig:mutual-information-0-slcp}. Further investigation of the hyperparameters on $-\hat{I}_{\bw}^{(0)}(\btheta; \bx)$ and $-\hat{I}_{\bw}^{(1)}(\btheta; \bx)$ can be found in Appendix~\ref{apndx:mutual-information}.

Our take-away is that $\gamma \in [0.1, 10]$ is a good range with negligible effects, although lower values tend to produce better estimates on complex problems at slower convergence. It seems plausible therefore that small $\gamma$ can act as a regularizer.  Outside of this range, learning can become unstable or slow. $\gamma=1$ has a good compromise on high convergence rate without sacrificing significant performance. Generally, we saw improved performance on the C2ST by increasing contrastive parameters $K$. Due to bootstrapping parameters from the batched $\btheta$, the maximum $K$ is $B/2$ without reusing any $\btheta$ to compare with $\bx$. (This is considering both terms in our loss function.)

In the end, we optimized our architecture based on the C2ST since that is the benchmark's metric, bias may be introduced due to the log partition function in $-\hat{I}_{\bw}^{(0)}(\btheta; \bx)$, and our naive computation of $-\hat{I}_{\bw}^{(1)}(\btheta; \bx)$ was too high variance to interpret in all cases. When the C2ST is not commutable, one could reduce the variance of $-\hat{I}_{\bw}^{(1)}(\btheta; \bx)$ with more samples. We did not take the maximum number of contrastive parameters, but rather a large value that suited our computation budget.

\paragraph{Benchmark}
We performed the benchmark tasks from \citet{sbibm} with \NREC. We made alternative architecture and hyperparameters choices based on the results of the last paragraph. We applied the largest number of computationally practical contrastive parameters, namely $K=99$, and set $\gamma = 1$. Since we identified potential architecture bottlenecks in Sections~\ref{sec:unlimited-joint} and \ref{sec:benchmark}, we used Large NN for our results. The averaged results are visible in Table~\ref{table:sbibm}. Sequential estimates are prepended with an \textsc{s}, all other estimates are amortized. We trained a five amortized estimators per task with five seeds, then we predicted the posterior for each of the ten observations $\bx_1, \ldots, \bx_{10}$. \citet{sbibm} trained an estimator for every $\bx$ with a new seed. The reported C2ST for our method is averaged across ten posteriors per seed with five seeds, i.e., fifty C2ST computations per task. The other methods are averaged across one posterior per seed with ten seeds, i.e., ten C2ST computations per task.

\NREC performed better than \NREB at all budgets and generally well among the amortized algorithms. The high performance at $10^{5}$ simulation budget implies that our design has high capacity and may scale well. Sequential methods naturally lead to high exactness since they are tailored to the specific observation $\bx_o$, but they come with drawbacks since they cannot practically perform empirical expected coverage tests \cite{hermans2021averting, miller2021truncated} nor bound the mutual information, as discussed in Section~\ref{sec:diagnostic}. Detailed results per task can be found in Appendix~\ref{apndx:experimental-details}.

\begin{wraptable}{R}{.5\textwidth}
    \centering
    \vspace{-12pt}
    \caption{Averaged C2ST for various budgets on the simulation-based inference benchmark \cite{sbibm}. An \textsc{s} implies a sequential i.e. non-amortized method.}
    \label{table:sbibm}
    \begin{tabular}{lrrr}
    \toprule
     & \multicolumn{3}{r}{C2ST} \\
    Simulation budget & $10^3$ & $10^4$ & $10^5$ \\
    Algorithm &  &  &  \\
    \midrule
    \NREC (ours) & 0.828 & 0.746 & 0.705 \\
    \REJABC & 0.965 & 0.920 & 0.871 \\
    \NLE & 0.826 & 0.753 & 0.723 \\
    \NPE & 0.838 & 0.736 & 0.654 \\
    \NRE (\NREB) & 0.867 & 0.811 & 0.762 \\
    \SMCABC & 0.948 & 0.873 & 0.816 \\
    \SNLE & 0.783 & 0.704 & 0.655 \\
    \SNPE & 0.796 & 0.677 & 0.615 \\
    \SNRE (\SNREB) & 0.788 & 0.703 & 0.610 \\
    \bottomrule
    \end{tabular}
\end{wraptable}

\section{Conclusion}
\label{sec:conclusion}
We introduced a generalization of \NREA and \NREB which discourages the \NREB specific $c_{\bw}(\bx)$ bias term and produces an unbiased estimate of the likelihood-to-evidence ratio at optimum. This property has implications for the importance sampling diagnostic, which is critical to practitioners.
We suggested using a variational bound on the mutual information as a model-selection metric that could replace the C2ST averaged over several observations. It is significantly more practical since it does not require access to the ground truth posterior; however, it remains a lower bound so we don't know the overall quality of our estimator. Additionally, both our proposed estimation techniques have either bias or high variance. Further discussions and derivations related to the mutual information are available in Appendix~\ref{apndx:mutual-information}.

In the context of \NREC, we found that increasing the number of contrastive parameters $K$ improved the C2ST in most cases. For fixed simulation data, low values of $\gamma$ slow convergence and could at as a regularizer in some complex problems, while high values of $\gamma$ increased the convergence rate and can introduced instability. We found that $\gamma \approx 1$ is a good setting.
These results indicate that better performance can be achieved by using \NREC with these hyperparameters than the other version of \NRE. When both training parameters $\btheta$ and simulation data $\bx$ are unlimited, the architecture size plays an important role in the quality of surrogate model.
We tried drawing contrastive parameters from the prior, a commonly available practical use-case, and found that it damped the effect of $K$ on the C2ST. According to the C2ST, for highest accuracy with a fixed budget set $K$ large, up to the mini-batch size divided by 2, and $\gamma \approx 1.0$ for the best convergence rate. When the prior is sampled for every mini-batch, applying a higher $K$ may be less valuable.

\paragraph{Broader Impact} The societal implications of \SBI are similar to other inference methods. The methods are primarily scientific and generally lead to interesting discoveries; however, one must be careful to use an accurate generative model and to carefully test empirical results. Model mismatch and untested inference can lead to incorrect conclusions. That is why we emphasize the importance of the diagnostics in our paper. This nuance can be missed by practitioners doing inference in any field; however, special care should be taken when producing inference results that may be used for making decisions in areas like predicting hidden variables that describe human behavior or determining what factors are responsible for climate change. This list is non-comprehensive and not specific to \SBI.

\section*{Errata and changes}
Previous versions of this paper had errors that we corrected here. There are additional changes.
\begin{enumerate}
    \item The code implementation incorrectly defined $\gamma$ and therefore the results from \NREC were not generated using the method we proposed in this paper. There were extraneous $K$ terms in the definition of $\gamma$ which had increasing effects as $K$ increased. All results have been reproduced in this version using the correct implementation.
    \item We did not satisfactorily emphasize the inherent bias in one of our estimates of the lower bound on the mutual information $I_{\bw}^{(0)}(\btheta; \bx)$. We have now introduced visualizations of $I_{\bw}^{(1)}(\btheta; \bx)$, which is unbiased, and explained the bias and variance of both estimators.
\end{enumerate}
These issues are present in versions one and two on arxiv and the published version in NeurIPS 2022. To distinguish this new version from previous ones we changed the name from \emph{Contrastive Neural Ratio Estimation} to \emph{Contrastive Neural Ratio Estimation for Simulation-based Inference}.

\begin{ack}
	We thank Noemi Anau Montel, James Alvey, Kosio Karchev for reading; Teodora Pandeva for mathematics consultation; Maud Canisius for design consultation; Marco Federici and Gilles Louppe for discussions. This work uses \texttt{numpy} \cite{harris2020array}, \texttt{scipy} \cite{2020SciPy-NMeth}, \texttt{seaborn} \cite{Waskom2021}, \texttt{matplotlib} \cite{Hunter:2007}, \texttt{pandas} \cite{reback2020pandas, mckinney-proc-scipy-2010}, \texttt{pytorch} \cite{pytorch}, and \texttt{jupyter} \cite{jupyter}. 

    We want to thank the DAS-5 computing cluster for access to their TitanX GPUs. DAS-5 is funded by the NWO/NCF (the Netherlands Organization for Scientific Research). We thank SURF (\href{www.surf.nl}{www.surf.nl}) for the support in using the Lisa Compute Cluster.

	Benjamin Kurt Miller is funded by the University of Amsterdam Faculty of Science (FNWI), Informatics Institute (IvI), and the Institute of Physics (IoP). 
	We received funding from the European Research Council (ERC) under the European Union’s Horizon 2020 research and innovation programme (Grant agreement No. 864035 -- UnDark).
\end{ack}

\bibliographystyle{abbrvnat}
\bibliography{bibliography}

\begin{thebibliography}{75}
\providecommand{\natexlab}[1]{#1}
\providecommand{\url}[1]{\texttt{#1}}
\expandafter\ifx\csname urlstyle\endcsname\relax
  \providecommand{\doi}[1]{doi: #1}\else
  \providecommand{\doi}{doi: \begingroup \urlstyle{rm}\Url}\fi

\bibitem[Alsing et~al.(2018)Alsing, Wandelt, and Feeney]{alsing2018massive}
J.~Alsing, B.~Wandelt, and S.~Feeney.
\newblock Massive optimal data compression and density estimation for scalable,
  likelihood-free inference in cosmology.
\newblock \emph{Monthly Notices of the Royal Astronomical Society},
  477\penalty0 (3):\penalty0 2874--2885, 2018.

\bibitem[Alsing et~al.(2019)Alsing, Charnock, Feeney, and
  Wandelt]{alsing2019fast}
J.~Alsing, T.~Charnock, S.~Feeney, and B.~Wandelt.
\newblock Fast likelihood-free cosmology with neural density estimators and
  active learning.
\newblock \emph{Monthly Notices of the Royal Astronomical Society},
  488\penalty0 (3):\penalty0 4440--4458, 2019.

\bibitem[Beaumont et~al.(2009)Beaumont, Cornuet, Marin, and
  Robert]{Beaumont2009-gl}
M.~A. Beaumont, J.-M. Cornuet, J.-M. Marin, and C.~P. Robert.
\newblock Adaptive approximate bayesian computation.
\newblock \emph{Biometrika}, 96\penalty0 (4):\penalty0 983--990, 2009.

\bibitem[Belghazi et~al.(2018)Belghazi, Baratin, Rajeswar, Ozair, Bengio,
  Courville, and Hjelm]{belghazi2018mine}
M.~I. Belghazi, A.~Baratin, S.~Rajeswar, S.~Ozair, Y.~Bengio, A.~Courville, and
  R.~D. Hjelm.
\newblock Mine: mutual information neural estimation.
\newblock \emph{arXiv preprint arXiv:1801.04062}, 2018.

\bibitem[Bishop(2006)]{bishop}
C.~M. Bishop.
\newblock Pattern recognition and machine learning (information science and
  statistics), 2006.

\bibitem[Blum and Fran{\c{c}}ois(2010)]{blum2010non}
M.~G. Blum and O.~Fran{\c{c}}ois.
\newblock Non-linear regression models for approximate bayesian computation.
\newblock \emph{Statistics and computing}, 20\penalty0 (1):\penalty0 63--73,
  2010.

\bibitem[Brehmer et~al.(2018)Brehmer, Cranmer, Louppe, and
  Pavez]{brehmer2018constraining}
J.~Brehmer, K.~Cranmer, G.~Louppe, and J.~Pavez.
\newblock Constraining effective field theories with machine learning.
\newblock \emph{Physical review letters}, 121\penalty0 (11):\penalty0 111801,
  2018.

\bibitem[Brehmer et~al.(2020)Brehmer, Louppe, Pavez, and
  Cranmer]{brehmer2020mining}
J.~Brehmer, G.~Louppe, J.~Pavez, and K.~Cranmer.
\newblock Mining gold from implicit models to improve likelihood-free
  inference.
\newblock \emph{Proceedings of the National Academy of Sciences}, 117\penalty0
  (10):\penalty0 5242--5249, 2020.

\bibitem[Chan et~al.(2018)Chan, Perrone, Spence, Jenkins, Mathieson, and
  Song]{chan2018likelihood}
J.~Chan, V.~Perrone, J.~Spence, P.~Jenkins, S.~Mathieson, and Y.~Song.
\newblock A likelihood-free inference framework for population genetic data
  using exchangeable neural networks.
\newblock \emph{Advances in neural information processing systems}, 31, 2018.

\bibitem[Cole et~al.(2021)Cole, Miller, Witte, Cai, Grootes, Nattino, and
  Weniger]{cole2021fast}
A.~Cole, B.~K. Miller, S.~J. Witte, M.~X. Cai, M.~W. Grootes, F.~Nattino, and
  C.~Weniger.
\newblock Fast and credible likelihood-free cosmology with truncated marginal
  neural ratio estimation.
\newblock \emph{arXiv preprint arXiv:2111.08030}, 2021.

\bibitem[Coogan et~al.(2020)Coogan, Karchev, and Weniger]{lensing}
A.~Coogan, K.~Karchev, and C.~Weniger.
\newblock Targeted likelihood-free inference of dark matter substructure in
  strongly-lensed galaxies.
\newblock \emph{arXiv preprint arXiv:2010.07032}, 2020.

\bibitem[Cranmer et~al.(2015)Cranmer, Pavez, and Louppe]{Cranmer2015}
K.~Cranmer, J.~Pavez, and G.~Louppe.
\newblock Approximating likelihood ratios with calibrated discriminative
  classifiers.
\newblock \emph{arXiv preprint arXiv:1506.02169}, 2015.

\bibitem[Cranmer et~al.(2020)Cranmer, Brehmer, and Louppe]{Cranmer2020}
K.~Cranmer, J.~Brehmer, and G.~Louppe.
\newblock The frontier of simulation-based inference.
\newblock \emph{Proc. Natl. Acad. Sci. U. S. A.}, May 2020.

\bibitem[Dalmasso et~al.(2020)Dalmasso, Izbicki, and
  Lee]{dalmasso2020confidence}
N.~Dalmasso, R.~Izbicki, and A.~Lee.
\newblock Confidence sets and hypothesis testing in a likelihood-free inference
  setting.
\newblock In \emph{International Conference on Machine Learning}, pages
  2323--2334. PMLR, 2020.

\bibitem[Drovandi et~al.(2018)Drovandi, Grazian, Mengersen, and
  Robert]{drovandi2018approximating}
C.~C. Drovandi, C.~Grazian, K.~Mengersen, and C.~Robert.
\newblock Approximating the likelihood in abc.
\newblock \emph{Handbook of approximate bayesian computation}, pages 321--368,
  2018.

\bibitem[Durkan et~al.(2020)Durkan, Murray, and Papamakarios]{Durkan2020}
C.~Durkan, I.~Murray, and G.~Papamakarios.
\newblock On contrastive learning for likelihood-free inference.
\newblock In \emph{International Conference on Machine Learning}, pages
  2771--2781. PMLR, 2020.

\bibitem[Friedman(2003)]{friedman2003multivariate}
J.~H. Friedman.
\newblock On multivariate goodness--of--fit and two--sample testing.
\newblock \emph{STATISTICAL PROBLEMS IN PARTICLE PHYSICS, ASTROPHYSICS AND
  COSMOLOGY}, page 311, 2003.

\bibitem[Gl{\"o}ckler et~al.(2021)Gl{\"o}ckler, Deistler, and
  Macke]{glockler2021variational}
M.~Gl{\"o}ckler, M.~Deistler, and J.~H. Macke.
\newblock Variational methods for simulation-based inference.
\newblock In \emph{International Conference on Learning Representations}, 2021.

\bibitem[Goodfellow et~al.(2020)Goodfellow, Pouget-Abadie, Mirza, Xu,
  Warde-Farley, Ozair, Courville, and Bengio]{goodfellow2014generative}
I.~Goodfellow, J.~Pouget-Abadie, M.~Mirza, B.~Xu, D.~Warde-Farley, S.~Ozair,
  A.~Courville, and Y.~Bengio.
\newblock Generative adversarial networks.
\newblock \emph{Communications of the ACM}, 63\penalty0 (11):\penalty0
  139--144, 2020.

\bibitem[Gratton(2017)]{gratton2017glass}
S.~Gratton.
\newblock Glass: A general likelihood approximate solution scheme.
\newblock \emph{arXiv preprint arXiv:1708.08479}, 2017.

\bibitem[Greenberg et~al.(2019)Greenberg, Nonnenmacher, and
  Macke]{greenberg2019automatic}
D.~Greenberg, M.~Nonnenmacher, and J.~Macke.
\newblock Automatic posterior transformation for likelihood-free inference.
\newblock In \emph{International Conference on Machine Learning}, pages
  2404--2414. PMLR, 2019.

\bibitem[Gutmann and Hyv{\"a}rinen(2010)]{gutmann2010noise}
M.~Gutmann and A.~Hyv{\"a}rinen.
\newblock Noise-contrastive estimation: A new estimation principle for
  unnormalized statistical models.
\newblock In \emph{Proceedings of the thirteenth international conference on
  artificial intelligence and statistics}, pages 297--304. JMLR Workshop and
  Conference Proceedings, 2010.

\bibitem[Gutmann and Hyv{\"a}rinen(2012)]{gutmann2012noise}
M.~U. Gutmann and A.~Hyv{\"a}rinen.
\newblock Noise-contrastive estimation of unnormalized statistical models, with
  applications to natural image statistics.
\newblock \emph{Journal of machine learning research}, 13\penalty0 (2), 2012.

\bibitem[Gutmann et~al.(2016)Gutmann, Corander, et~al.]{gutmann2016bayesian}
M.~U. Gutmann, J.~Corander, et~al.
\newblock Bayesian optimization for likelihood-free inference of
  simulator-based statistical models.
\newblock \emph{Journal of Machine Learning Research}, 2016.

\bibitem[Gutmann et~al.(2022)Gutmann, Kleinegesse, and
  Rhodes]{gutmann2022statistical}
M.~U. Gutmann, S.~Kleinegesse, and B.~Rhodes.
\newblock Statistical applications of contrastive learning.
\newblock \emph{arXiv preprint arXiv:2204.13999}, 2022.

\bibitem[Harris et~al.(2020)Harris, Millman, van~der Walt, Gommers, Virtanen,
  Cournapeau, Wieser, Taylor, Berg, Smith, Kern, Picus, Hoyer, van Kerkwijk,
  Brett, Haldane, del R{'{\i}}o, Wiebe, Peterson, G{'{e}}rard-Marchant,
  Sheppard, Reddy, Weckesser, Abbasi, Gohlke, and Oliphant]{harris2020array}
C.~R. Harris, K.~J. Millman, S.~J. van~der Walt, R.~Gommers, P.~Virtanen,
  D.~Cournapeau, E.~Wieser, J.~Taylor, S.~Berg, N.~J. Smith, R.~Kern, M.~Picus,
  S.~Hoyer, M.~H. van Kerkwijk, M.~Brett, A.~Haldane, J.~F. del R{'{\i}}o,
  M.~Wiebe, P.~Peterson, P.~G{'{e}}rard-Marchant, K.~Sheppard, T.~Reddy,
  W.~Weckesser, H.~Abbasi, C.~Gohlke, and T.~E. Oliphant.
\newblock Array programming with {NumPy}.
\newblock \emph{Nature}, 585\penalty0 (7825):\penalty0 357--362, Sept. 2020.
\newblock \doi{10.1038/s41586-020-2649-2}.
\newblock URL \url{https://doi.org/10.1038/s41586-020-2649-2}.

\bibitem[Hastie et~al.(2009)Hastie, Tibshirani, Friedman, and
  Friedman]{hastie2009elements}
T.~Hastie, R.~Tibshirani, J.~H. Friedman, and J.~H. Friedman.
\newblock \emph{The elements of statistical learning: data mining, inference,
  and prediction}, volume~2.
\newblock Springer, 2009.

\bibitem[He et~al.(2016)He, Zhang, Ren, and Sun]{resnet}
K.~He, X.~Zhang, S.~Ren, and J.~Sun.
\newblock Deep residual learning for image recognition.
\newblock In \emph{2016 IEEE Conference on Computer Vision and Pattern
  Recognition (CVPR)}, pages 770--778, 2016.
\newblock \doi{10.1109/CVPR.2016.90}.

\bibitem[Hermans et~al.(2020{\natexlab{a}})Hermans, Banik, Weniger, Bertone,
  and Louppe]{hermans2020towards}
J.~Hermans, N.~Banik, C.~Weniger, G.~Bertone, and G.~Louppe.
\newblock Towards constraining warm dark matter with stellar streams through
  neural simulation-based inference.
\newblock \emph{arXiv preprint arXiv:2011.14923}, 2020{\natexlab{a}}.

\bibitem[Hermans et~al.(2020{\natexlab{b}})Hermans, Begy, and
  Louppe]{Hermans2019}
J.~Hermans, V.~Begy, and G.~Louppe.
\newblock Likelihood-free mcmc with amortized approximate ratio estimators.
\newblock In \emph{International Conference on Machine Learning}, pages
  4239--4248. PMLR, 2020{\natexlab{b}}.

\bibitem[Hermans et~al.(2021)Hermans, Delaunoy, Rozet, Wehenkel, and
  Louppe]{hermans2021averting}
J.~Hermans, A.~Delaunoy, F.~Rozet, A.~Wehenkel, and G.~Louppe.
\newblock Averting a crisis in simulation-based inference.
\newblock \emph{arXiv preprint arXiv:2110.06581}, 2021.

\bibitem[Hunter(2007)]{Hunter:2007}
J.~D. Hunter.
\newblock Matplotlib: A 2d graphics environment.
\newblock \emph{Computing in Science \& Engineering}, 9\penalty0 (3):\penalty0
  90--95, 2007.
\newblock \doi{10.1109/MCSE.2007.55}.

\bibitem[Ioffe and Szegedy(2015)]{ioffe2015batch}
S.~Ioffe and C.~Szegedy.
\newblock Batch normalization: Accelerating deep network training by reducing
  internal covariate shift.
\newblock In \emph{International conference on machine learning}, pages
  448--456. PMLR, 2015.

\bibitem[Izbicki et~al.(2014)Izbicki, Lee, and Schafer]{izbicki2014high}
R.~Izbicki, A.~Lee, and C.~Schafer.
\newblock High-dimensional density ratio estimation with extensions to
  approximate likelihood computation.
\newblock In \emph{Artificial intelligence and statistics}, pages 420--429.
  PMLR, 2014.

\bibitem[J{\"a}rvenp{\"a}{\"a} et~al.(2021)J{\"a}rvenp{\"a}{\"a}, Gutmann,
  Vehtari, and Marttinen]{jarvenpaa2021parallel}
M.~J{\"a}rvenp{\"a}{\"a}, M.~U. Gutmann, A.~Vehtari, and P.~Marttinen.
\newblock Parallel gaussian process surrogate bayesian inference with noisy
  likelihood evaluations.
\newblock \emph{Bayesian Analysis}, 16\penalty0 (1):\penalty0 147--178, 2021.

\bibitem[Jeffrey and Wandelt(2020)]{jeffrey2020solving}
N.~Jeffrey and B.~D. Wandelt.
\newblock Solving high-dimensional parameter inference: marginal posterior
  densities \& moment networks.
\newblock \emph{arXiv preprint arXiv:2011.05991}, 2020.

\bibitem[Kingma and Ba(2015)]{kingma2015adam}
D.~P. Kingma and J.~L. Ba.
\newblock Adam: A method for stochastic gradient descent.
\newblock In \emph{ICLR: International Conference on Learning Representations},
  pages 1--15, 2015.

\bibitem[Kluyver et~al.(2016)Kluyver, Ragan-Kelley, P{\'e}rez, Granger,
  Bussonnier, Frederic, Kelley, Hamrick, Grout, Corlay, Ivanov, Avila, Abdalla,
  Willing, and development team]{jupyter}
T.~Kluyver, B.~Ragan-Kelley, F.~P{\'e}rez, B.~Granger, M.~Bussonnier,
  J.~Frederic, K.~Kelley, J.~Hamrick, J.~Grout, S.~Corlay, P.~Ivanov, D.~Avila,
  S.~Abdalla, C.~Willing, and J.~development team.
\newblock Jupyter notebooks - a publishing format for reproducible
  computational workflows.
\newblock In F.~Loizides and B.~Scmidt, editors, \emph{Positioning and Power in
  Academic Publishing: Players, Agents and Agendas}, pages 87--90, Netherlands,
  2016. IOS Press.
\newblock URL \url{https://eprints.soton.ac.uk/403913/}.

\bibitem[Lacoste et~al.(2019)Lacoste, Luccioni, Schmidt, and
  Dandres]{lacoste2019quantifying}
A.~Lacoste, A.~Luccioni, V.~Schmidt, and T.~Dandres.
\newblock Quantifying the carbon emissions of machine learning.
\newblock \emph{arXiv preprint arXiv:1910.09700}, 2019.

\bibitem[Lehmann et~al.(2005)Lehmann, Romano, and Casella]{lehmann2005testing}
E.~L. Lehmann, J.~P. Romano, and G.~Casella.
\newblock \emph{Testing statistical hypotheses}, volume~3.
\newblock Springer, 2005.

\bibitem[Lopez-Paz and Oquab(2017)]{lopez2017revisiting}
D.~Lopez-Paz and M.~Oquab.
\newblock Revisiting classifier two-sample tests.
\newblock In \emph{International Conference on Learning Representations}, 2017.

\bibitem[Lueckmann et~al.(2017)Lueckmann, Gon{\c{c}}alves, Bassetto, {\"O}cal,
  Nonnenmacher, and Macke]{lueckmann2017flexible}
J.-M. Lueckmann, P.~J. Gon{\c{c}}alves, G.~Bassetto, K.~{\"O}cal,
  M.~Nonnenmacher, and J.~H. Macke.
\newblock Flexible statistical inference for mechanistic models of neural
  dynamics.
\newblock In \emph{Proceedings of the 31st International Conference on Neural
  Information Processing Systems}, pages 1289--1299, 2017.

\bibitem[Lueckmann et~al.(2019)Lueckmann, Bassetto, Karaletsos, and
  Macke]{lueckmann2019likelihood}
J.-M. Lueckmann, G.~Bassetto, T.~Karaletsos, and J.~H. Macke.
\newblock Likelihood-free inference with emulator networks.
\newblock In \emph{Symposium on Advances in Approximate Bayesian Inference},
  pages 32--53. PMLR, 2019.

\bibitem[Lueckmann et~al.(2021)Lueckmann, Boelts, Greenberg, Goncalves, and
  Macke]{sbibm}
J.-M. Lueckmann, J.~Boelts, D.~Greenberg, P.~Goncalves, and J.~Macke.
\newblock Benchmarking simulation-based inference.
\newblock In A.~Banerjee and K.~Fukumizu, editors, \emph{Proceedings of The
  24th International Conference on Artificial Intelligence and Statistics},
  volume 130 of \emph{Proceedings of Machine Learning Research}, pages
  343--351. PMLR, 13--15 Apr 2021.
\newblock URL \url{http://proceedings.mlr.press/v130/lueckmann21a.html}.

\bibitem[Ma and Collins(2018)]{ma2018noise}
Z.~Ma and M.~Collins.
\newblock Noise contrastive estimation and negative sampling for conditional
  models: Consistency and statistical efficiency.
\newblock In \emph{Proceedings of the 2018 Conference on Empirical Methods in
  Natural Language Processing}, pages 3698--3707, 2018.

\bibitem[Mikolov et~al.(2013)Mikolov, Sutskever, Chen, Corrado, and
  Dean]{mikolov2013distributed}
T.~Mikolov, I.~Sutskever, K.~Chen, G.~S. Corrado, and J.~Dean.
\newblock Distributed representations of words and phrases and their
  compositionality.
\newblock \emph{Advances in neural information processing systems}, 26, 2013.

\bibitem[Miller et~al.(2020)Miller, Cole, Louppe, and
  Weniger]{miller2020simulation}
B.~K. Miller, A.~Cole, G.~Louppe, and C.~Weniger.
\newblock Simulation-efficient marginal posterior estimation with swyft: stop
  wasting your precious time.
\newblock \emph{arXiv preprint arXiv:2011.13951}, 2020.

\bibitem[Miller et~al.(2021)Miller, Cole, Forr{\'e}, Louppe, and
  Weniger]{miller2021truncated}
B.~K. Miller, A.~Cole, P.~Forr{\'e}, G.~Louppe, and C.~Weniger.
\newblock Truncated marginal neural ratio estimation.
\newblock \emph{Advances in Neural Information Processing Systems},
  34:\penalty0 129--143, 2021.

\bibitem[Miller et~al.(2023)Miller, Federici, Weniger, and
  Forr{\'e}]{miller2023simulation}
B.~K. Miller, M.~Federici, C.~Weniger, and P.~Forr{\'e}.
\newblock Simulation-based inference with the generalized kullback-leibler
  divergence.
\newblock In \emph{1st Workshop on the Synergy of Scientific and Machine
  Learning Modeling@ ICML2023}, 2023.

\bibitem[Mnih and Teh(2012)]{mnih2012fast}
A.~Mnih and Y.~W. Teh.
\newblock A fast and simple algorithm for training neural probabilistic
  language models.
\newblock In \emph{Proceedings of the 29th International Coference on
  International Conference on Machine Learning}, pages 419--426, 2012.

\bibitem[Mohamed and Lakshminarayanan(2016)]{Mohamed2016}
S.~Mohamed and B.~Lakshminarayanan.
\newblock Learning in implicit generative models.
\newblock \emph{arXiv preprint arXiv:1610.03483}, 2016.

\bibitem[Neal(2003)]{neal2003slice}
R.~M. Neal.
\newblock Slice sampling.
\newblock \emph{Annals of statistics}, pages 705--741, 2003.

\bibitem[pandas~development team(2020)]{reback2020pandas}
T.~pandas~development team.
\newblock pandas-dev/pandas: Pandas, Feb. 2020.
\newblock URL \url{https://doi.org/10.5281/zenodo.3509134}.

\bibitem[Papamakarios and Murray(2016)]{papamakarios2016fast}
G.~Papamakarios and I.~Murray.
\newblock Fast $\varepsilon$-free inference of simulation models with bayesian
  conditional density estimation.
\newblock \emph{Advances in neural information processing systems}, 29, 2016.

\bibitem[Papamakarios et~al.(2017)Papamakarios, Pavlakou, and
  Murray]{papamakarios2017masked}
G.~Papamakarios, T.~Pavlakou, and I.~Murray.
\newblock Masked autoregressive flow for density estimation.
\newblock In \emph{Proceedings of the 31st International Conference on Neural
  Information Processing Systems}, pages 2335--2344, 2017.

\bibitem[Papamakarios et~al.(2019{\natexlab{a}})Papamakarios, Nalisnick,
  Rezende, Mohamed, and Lakshminarayanan]{papamakarios2019normalizing}
G.~Papamakarios, E.~Nalisnick, D.~J. Rezende, S.~Mohamed, and
  B.~Lakshminarayanan.
\newblock Normalizing flows for probabilistic modeling and inference.
\newblock \emph{arXiv preprint arXiv:1912.02762}, 2019{\natexlab{a}}.

\bibitem[Papamakarios et~al.(2019{\natexlab{b}})Papamakarios, Sterratt, and
  Murray]{papamakarios2019sequential}
G.~Papamakarios, D.~Sterratt, and I.~Murray.
\newblock Sequential neural likelihood: Fast likelihood-free inference with
  autoregressive flows.
\newblock In \emph{The 22nd International Conference on Artificial Intelligence
  and Statistics}, pages 837--848. PMLR, 2019{\natexlab{b}}.

\bibitem[Paszke et~al.(2019)Paszke, Gross, Massa, Lerer, Bradbury, Chanan,
  Killeen, Lin, Gimelshein, Antiga, Desmaison, Kopf, Yang, DeVito, Raison,
  Tejani, Chilamkurthy, Steiner, Fang, Bai, and Chintala]{pytorch}
A.~Paszke, S.~Gross, F.~Massa, A.~Lerer, J.~Bradbury, G.~Chanan, T.~Killeen,
  Z.~Lin, N.~Gimelshein, L.~Antiga, A.~Desmaison, A.~Kopf, E.~Yang, Z.~DeVito,
  M.~Raison, A.~Tejani, S.~Chilamkurthy, B.~Steiner, L.~Fang, J.~Bai, and
  S.~Chintala.
\newblock Pytorch: An imperative style, high-performance deep learning library.
\newblock In H.~Wallach, H.~Larochelle, A.~Beygelzimer, F.~d~Alch\'{e}-Buc,
  E.~Fox, and R.~Garnett, editors, \emph{Advances in Neural Information
  Processing Systems 32}, pages 8024--8035. Curran Associates, Inc., 2019.
\newblock URL
  \url{http://papers.neurips.cc/paper/9015-pytorch-an-imperative-style-high-performance-deep-learning-library.pdf}.

\bibitem[Poole et~al.(2019)Poole, Ozair, Van Den~Oord, Alemi, and
  Tucker]{poole2019variational}
B.~Poole, S.~Ozair, A.~Van Den~Oord, A.~Alemi, and G.~Tucker.
\newblock On variational bounds of mutual information.
\newblock In \emph{International Conference on Machine Learning}, pages
  5171--5180. PMLR, 2019.

\bibitem[Prangle(2019)]{prangle2019distilling}
D.~Prangle.
\newblock Distilling importance sampling.
\newblock \emph{arXiv preprint arXiv:1910.03632}, 2019.

\bibitem[Ramesh et~al.(2021)Ramesh, Lueckmann, Boelts, Tejero-Cantero,
  Greenberg, Goncalves, and Macke]{ramesh2021gatsbi}
P.~Ramesh, J.-M. Lueckmann, J.~Boelts, {\'A}.~Tejero-Cantero, D.~S. Greenberg,
  P.~J. Goncalves, and J.~H. Macke.
\newblock Gatsbi: Generative adversarial training for simulation-based
  inference.
\newblock In \emph{International Conference on Learning Representations}, 2021.

\bibitem[Rodrigues et~al.(2020)Rodrigues, Nott, and
  Sisson]{rodrigues2020likelihood}
G.~Rodrigues, D.~J. Nott, and S.~A. Sisson.
\newblock Likelihood-free approximate gibbs sampling.
\newblock \emph{Statistics and Computing}, 30\penalty0 (4):\penalty0
  1057--1073, 2020.

\bibitem[Shimodaira(2000)]{shimodaira2000improving}
H.~Shimodaira.
\newblock Improving predictive inference under covariate shift by weighting the
  log-likelihood function.
\newblock \emph{Journal of statistical planning and inference}, 90\penalty0
  (2):\penalty0 227--244, 2000.

\bibitem[Simola et~al.(2020)Simola, Corander, and Picchini]{simola2020adaptive}
U.~Simola, J.~Corander, and U.~Picchini.
\newblock Adaptive mcmc for synthetic likelihoods and correlated synthetic
  likelihoods.
\newblock \emph{Bayesian analysis}, 2020.

\bibitem[Sisson et~al.(2018)Sisson, Fan, and Beaumont]{sisson2018handbook}
S.~A. Sisson, Y.~Fan, and M.~Beaumont.
\newblock \emph{Handbook of approximate Bayesian computation}.
\newblock CRC Press, 2018.

\bibitem[Skilling(2006)]{Skilling2006}
J.~Skilling.
\newblock Nested sampling for general bayesian computation.
\newblock \emph{Bayesian Anal.}, 1\penalty0 (4):\penalty0 833--859, Dec. 2006.

\bibitem[Sugiyama et~al.(2012)Sugiyama, Suzuki, and
  Kanamori]{sugiyama2012density}
M.~Sugiyama, T.~Suzuki, and T.~Kanamori.
\newblock \emph{Density ratio estimation in machine learning}.
\newblock Cambridge University Press, 2012.

\bibitem[Talts et~al.(2018)Talts, Betancourt, Simpson, Vehtari, and
  Gelman]{talts2018validating}
S.~Talts, M.~Betancourt, D.~Simpson, A.~Vehtari, and A.~Gelman.
\newblock Validating bayesian inference algorithms with simulation-based
  calibration.
\newblock \emph{arXiv preprint arXiv:1804.06788}, 2018.

\bibitem[Tejero-Cantero et~al.(2020)Tejero-Cantero, Boelts, Deistler,
  Lueckmann, Durkan, Gonçalves, Greenberg, and Macke]{sbi}
A.~Tejero-Cantero, J.~Boelts, M.~Deistler, J.-M. Lueckmann, C.~Durkan, P.~J.
  Gonçalves, D.~S. Greenberg, and J.~H. Macke.
\newblock sbi: A toolkit for simulation-based inference.
\newblock \emph{Journal of Open Source Software}, 5\penalty0 (52):\penalty0
  2505, 2020.
\newblock \doi{10.21105/joss.02505}.
\newblock URL \url{https://doi.org/10.21105/joss.02505}.

\bibitem[Thomas et~al.(2016)Thomas, Dutta, Corander, Kaski, Gutmann,
  et~al.]{thomas2016likelihood}
O.~Thomas, R.~Dutta, J.~Corander, S.~Kaski, M.~U. Gutmann, et~al.
\newblock Likelihood-free inference by ratio estimation.
\newblock \emph{Bayesian Analysis}, 2016.

\bibitem[Toni et~al.(2009)Toni, Welch, Strelkowa, Ipsen, and
  Stumpf]{Toni2009-fd}
T.~Toni, D.~Welch, N.~Strelkowa, A.~Ipsen, and M.~P.~H. Stumpf.
\newblock Approximate bayesian computation scheme for parameter inference and
  model selection in dynamical systems.
\newblock \emph{J. R. Soc. Interface}, 6\penalty0 (31):\penalty0 187--202, Feb.
  2009.

\bibitem[Van~den Oord et~al.(2018)Van~den Oord, Li, and
  Vinyals]{van2018representation}
A.~Van~den Oord, Y.~Li, and O.~Vinyals.
\newblock Representation learning with contrastive predictive coding.
\newblock \emph{arXiv e-prints}, pages arXiv--1807, 2018.

\bibitem[Virtanen et~al.(2020)Virtanen, Gommers, Oliphant, Haberland, Reddy,
  Cournapeau, Burovski, Peterson, Weckesser, Bright, {van der Walt}, Brett,
  Wilson, Millman, Mayorov, Nelson, Jones, Kern, Larson, Carey, Polat, Feng,
  Moore, {VanderPlas}, Laxalde, Perktold, Cimrman, Henriksen, Quintero, Harris,
  Archibald, Ribeiro, Pedregosa, {van Mulbregt}, and {SciPy 1.0
  Contributors}]{2020SciPy-NMeth}
P.~Virtanen, R.~Gommers, T.~E. Oliphant, M.~Haberland, T.~Reddy, D.~Cournapeau,
  E.~Burovski, P.~Peterson, W.~Weckesser, J.~Bright, S.~J. {van der Walt},
  M.~Brett, J.~Wilson, K.~J. Millman, N.~Mayorov, A.~R.~J. Nelson, E.~Jones,
  R.~Kern, E.~Larson, C.~J. Carey, {\.I}.~Polat, Y.~Feng, E.~W. Moore,
  J.~{VanderPlas}, D.~Laxalde, J.~Perktold, R.~Cimrman, I.~Henriksen, E.~A.
  Quintero, C.~R. Harris, A.~M. Archibald, A.~H. Ribeiro, F.~Pedregosa, P.~{van
  Mulbregt}, and {SciPy 1.0 Contributors}.
\newblock {{SciPy} 1.0: Fundamental Algorithms for Scientific Computing in
  Python}.
\newblock \emph{Nature Methods}, 17:\penalty0 261--272, 2020.
\newblock \doi{10.1038/s41592-019-0686-2}.

\bibitem[Waskom(2021)]{Waskom2021}
M.~L. Waskom.
\newblock seaborn: statistical data visualization.
\newblock \emph{Journal of Open Source Software}, 6\penalty0 (60):\penalty0
  3021, 2021.
\newblock \doi{10.21105/joss.03021}.
\newblock URL \url{https://doi.org/10.21105/joss.03021}.

\bibitem[{W}es {M}c{K}inney(2010)]{mckinney-proc-scipy-2010}
{W}es {M}c{K}inney.
\newblock {D}ata {S}tructures for {S}tatistical {C}omputing in {P}ython.
\newblock In {S}t\'efan van~der {W}alt and {J}arrod {M}illman, editors,
  \emph{{P}roceedings of the 9th {P}ython in {S}cience {C}onference}, pages 56
  -- 61, 2010.
\newblock \doi{10.25080/Majora-92bf1922-00a}.

\end{thebibliography}

\clearpage

\section*{Checklist}

The checklist follows the references.  Please
read the checklist guidelines carefully for information on how to answer these
questions.  For each question, change the default \answerTODO{} to \answerYes{},
\answerNo{}, or \answerNA{}.  You are strongly encouraged to include a {\bf
	justification to your answer}, either by referencing the appropriate section of
your paper or providing a brief inline description.  For example:
\begin{itemize}
	\item Did you include the license to the code and datasets? \answerYes{See Section~.}
	\item Did you include the license to the code and datasets? \answerNo{The code and the data are proprietary.}
	\item Did you include the license to the code and datasets? \answerNA{}
\end{itemize}
Please do not modify the questions and only use the provided macros for your
answers.  Note that the Checklist section does not count towards the page
limit.  In your paper, please delete this instructions block and only keep the
Checklist section heading above along with the questions/answers below.

\begin{enumerate}

	\item For all authors...
	\begin{enumerate}
		\item Do the main claims made in the abstract and introduction accurately reflect the paper's contributions and scope?
		\answerYes{We introduce the generalization in Section~\ref{sec:nrec}. We show that it has the properties we claim in Appendix~\ref{apndx:proof}. We perform the experiments in Section~\ref{sec:experiments} and make a hyperparameter recommendation in Section~\ref{sec:conclusion}}
		\item Did you describe the limitations of your work?
		\answerYes{We discussed the behavior of ours and other algorithms in the finite sample setting, see Section~\ref{sec:nrec} and limitations of existing to test the convergence of our method, see Section~\ref{sec:diagnostic}.}
		\item Did you discuss any potential negative societal impacts of your work?
		\answerYes{We briefly mentioned them in Section~\ref{sec:conclusion}.}
		\item Have you read the ethics review guidelines and ensured that your paper conforms to them?
		\answerYes{This work applies toy simulation data thus no ethics concerns were raised by the paper. Other matters of ethics during experimentation and writing conformed.}
	\end{enumerate}

	\item If you are including theoretical results...
	\begin{enumerate}
		\item Did you state the full set of assumptions of all theoretical results?
		\answerYes{These are laid out in Appendix~\ref{apndx:proof}.}
		\item Did you include complete proofs of all theoretical results?
		\answerYes{We have only one real theoretical result and that is proven in Lemma~\ref{lem:optimal=log-ratio}. Anything else follows algebraically and is shown in detail.}
	\end{enumerate}

	\item If you ran experiments...
	\begin{enumerate}
		\item Did you include the code, data, and instructions needed to reproduce the main experimental results (either in the supplemental material or as a URL)?
		\answerYes{The code will be made available via a link in the supplemental material. It is a python package with example calls and self-explanatory installation instructions. The details will be shown in Section~\ref{apndx:experimental-details}. Results will be released upon acceptance.}
		\item Did you specify all the training details (e.g., data splits, hyperparameters, how they were chosen)?
		\answerYes{Additionally to the hyperparameter search which is the primary experimental result, seen in Section~\ref{sec:experiments}, we included more details in Appendix~\ref{apndx:experimental-details}.}
		\item Did you report error bars (e.g., with respect to the random seed after running experiments multiple times)?
		\answerYes{The hyperparameter searches indeed included error bars in Section~\ref{sec:experiments}. We did not include error bars in the computation of the diagnostic since we emphasize that it is made possible by our method, not its statistical properties. The details of the benchmark, including some uncertainty in the last experimental section are shown in Appendix~\ref{apndx:experimental-details}.}
		\item Did you include the total amount of compute and the type of resources used (e.g., type of GPUs, internal cluster, or cloud provider)?
		\answerYes{This is discussed in Appendix~\ref{apndx:experimental-details}.}
	\end{enumerate}

	\item If you are using existing assets (e.g., code, data, models) or curating/releasing new assets...
	\begin{enumerate}
		\item If your work uses existing assets, did you cite the creators?
		\answerYes{We cited the benchmark \cite{sbibm} several times in the paper. Also the sbi package \cite{sbi}}
		\item Did you mention the license of the assets?
		\answerYes{The license is mentioned in Appendix~\ref{apndx:other-sbi}}
		\item Did you include any new assets either in the supplemental material or as a URL?
		\answerYes{I will link to my code in the supplemental material as discussed above. Upon acceptance results will be made available.}
		\item Did you discuss whether and how consent was obtained from people whose data you're using/curating?
		\answerNA{The publicly available data was used in accordance with the aforementioned license in \cite{sbibm}. There was no need to get consent or mention it in the paper.}
		\item Did you discuss whether the data you are using/curating contains personally identifiable information or offensive content?
		\answerNA{The data is generated by toy mathematical models which abstractly relate parameters to data. It is highly unlikely any culture would find them offensive.}
	\end{enumerate}

	\item If you used crowdsourcing or conducted research with human subjects...
	\begin{enumerate}
		\item Did you include the full text of instructions given to participants and screenshots, if applicable?
		\answerNA{}
		\item Did you describe any potential participant risks, with links to Institutional Review Board (IRB) approvals, if applicable?
		\answerNA{}
		\item Did you include the estimated hourly wage paid to participants and the total amount spent on participant compensation?
		\answerNA{}
	\end{enumerate}

\end{enumerate}

\clearpage

\appendix

\section{Relationship to other \SBI methods}
\label{apndx:other-sbi}

\paragraph{Sampling in \SBI}
All \NRE (and \SBI) algorithms require samples from the joint distribution $p(\btheta, \bx)$. \NRE additionally requires samples from the product of marginals $p(\btheta)p(\bx)$. Sampling the joint with a simulator requires drawing from the prior $\btheta \sim p(\btheta)$ then passing that parameter into the simulator to produce $\bx \sim p(\bx \mid \btheta)$. Sampling the product of marginals is simple, just take another sample from the prior $\btheta' \sim p(\btheta)$ and pair it with our simulation from before. Then we have $(\btheta, \bx) \sim p(\btheta, \bx)$ and $(\btheta', \bx) \sim p(\btheta)p(\bx)$. In practice, we refer to this operation as a bootstrap within a mini-batch where we take $\btheta'$ from other parameter-simulation pairs and reuse them to create samples drawn from the product of marginals $p(\btheta)p(\bx)$. \NREC sometimes requires more $\btheta$ samples, often represented as $\bTheta$. These can be generated by merely concatenating several samples from $p(\btheta)$. Some \SBI methods, e.g., sequential methods, sometimes replace the prior with a proposal distribution $\ptilde(\btheta)$.

\paragraph{Amortized and sequential \SBI}
Recently, significant progress has been made in \SBI, especially with so-called \emph{sequential} methods that use active learning to draw samples from the posterior for a fixed observation-of-interest $\bxo$ \cite{gutmann2016bayesian, papamakarios2016fast, thomas2016likelihood, lueckmann2017flexible, gratton2017glass, papamakarios2019sequential, greenberg2019automatic, prangle2019distilling, Durkan2020, simola2020adaptive, rodrigues2020likelihood, jarvenpaa2021parallel, ramesh2021gatsbi, glockler2021variational}.
\emph{Amortized} \SBI algorithms, that can draw samples from the posterior for arbitrary observation $\bx$, have also enjoyed attentive development \cite{chan2018likelihood, Hermans2019, jeffrey2020solving, brehmer2020mining, miller2020simulation, miller2021truncated}. \citet{hermans2021averting} and \citet{miller2021truncated} argue that their intrinsic empirical testability makes amortized methods better applicable to the scientific use-case despite their inherently higher training expense. The last pillar of development has been into assessment methods that determine the reliability of approximate inference results. In the machine learning community, the focus has been on evaluating the exactness of estimates for tractable problems \cite{sbibm}. 
Evaluation methods which apply in the practical case where no tractable posterior is available are under development \cite{Cranmer2015, brehmer2018constraining, talts2018validating, dalmasso2020confidence, hermans2020towards}. We make a contribution to this effort in Appendix~\ref{apndx:mutual-information}.

\paragraph{Contrastive learning, \NREB, and \NPE}
We call our method Contrastive Neural Ratio Estimation because it the classifier is trained to identify which pairs of $\btheta$ and $\bx$ should be paired together. \citet{gutmann2022statistical} created an overview of contrastive learning for statistical problems generally. Specific connections are in the loss functions with \NREA closely corresponding to noise-contrastive estimation (NCE) \cite{gutmann2010noise, gutmann2012noise} and \NREB with RankingNCE, InfoNCE, and related \cite{mikolov2013distributed, ma2018noise, mnih2012fast, van2018representation}.

A core aspect of our paper focus on the effects of the \NREB ratio estimate biased by $c_{\bw}(\bx)$ at optimum. \citet{ma2018noise} also investigate the effects on the partition function when applying a binary or multi-class loss variant for estimating conditional energy-based models. Due to the similarity in the loss functions, they exhibit a similar bias in their partition function. In order to correct this bias they estimate $c_{\bw}(\bx)$ directly. We did not attempt to do this for our problem, so we do not have intuition about the effectiveness of such an approach to the likelihood-based diagnostic in \NRE. Although, since $c_{\bw}(\bx)$ is completely unconstrained, it may be quite difficult to estimate. We believe that this would be an alternative direction for future work.

\citet{Durkan2020} emphasize the connection between \NRE and contrastive learning in their paper which created a framework such that Neural Posterior Estimation (\NPE) and \NREB can be trained using the same loss function. The addition of an independently drawn $(\bTheta, \bx)$ set means that the \NREC framework is not trivially applicable for computing the normalizing constant on atoms necessary for their sequential version of \NPE.

\paragraph{Software}
Our experiments used \texttt{sbi} \cite{sbi}, which is released under an AGPL-3.0 license, and \texttt{sbibm} \cite{sbibm}, which is released under an MIT license. They implement various neural \SBI algorithms and benchmark problems respectively.

\clearpage

\subsection{Full derivation of other \NRE methods using our framework}

Here we present a derivation of the previous works \NREA \cite{Hermans2019} and \NREB \cite{Hermans2019} in our framwork.

\paragraph{\NREA} To estimate $r(\bx \mid \btheta)$, \citet{Hermans2019} introduce an indicator variable $y$ which switches between dependently and independently drawn samples. We have conditional probability

\begin{equation}
	p_{\NREA}(\btheta, \bx \mid y = k) = 
	\begin{cases}
		p(\btheta) p(\bx) & k = 0 \\
		p(\btheta, \bx) & k = 1
	\end{cases}.
\end{equation}
 
Each class' marginal probability is set equally, $p(y=0) = p(y=1)$. Dropping the \NREA subscript, the probability that $(\btheta, \bx)$ was drawn jointly is encoded in the another conditional probability
 
\begin{equation}
\begin{aligned}
    p(y=1 \mid \btheta, \bx) 
    &= \frac{p(\btheta, \bx \mid y=1) p(y=1)}{p(\btheta, \bx \mid y=1) p(y=1) + p(\btheta, \bx \mid y=0) p(y=0)} \\
    &= \frac{p(\btheta, \bx)}{p(\btheta, \bx) + p(\btheta) p(\bx)} 
    = \frac{r(\bx \mid \btheta)}{1 + r(\bx \mid \btheta)}.
\end{aligned}
\end{equation}

\NREA estimates $\log \rhat(\bx \mid \btheta)$ with neural network $f_{\bw}$ with weights $\bw$. Training is done by minimizing the binary cross-entropy of $q_{\bw}(y = 1 \mid \btheta, \bx) \coloneqq \sigma \circ f_{\bw}(\btheta, \bx)$ relative to $p(y, \btheta, \bx)$. For $B$ samples,
\begin{equation}
    \bw = \argmin_{\bw} \left[- \frac{1}{B} \sum_{b=1}^{B} \log \left(
        1 - \sigma \circ f_{\bw}(\btheta^{(b)}, \bx^{(b)})
    \right) 
    - \frac{1}{B}  \sum_{b'=1}^{B}\log \left(
        \sigma \circ f_{\bw}(\btheta^{(b')}, \bx^{(b')})
    \right) \right]
\end{equation}
 
where $\btheta^{(b)}, \bx^{(b)} \sim p(\btheta) p(\bx)$ and $\btheta^{(b')}, \bx^{(b')} \sim p(\btheta, \bx)$. In practice, $\btheta^{(b')}$ is bootstrapped from within the mini-batch to produce $\btheta^{(b)}$. \NREA's ratio estimate converges to ${f_{\bw} = \log \frac{p(\bx \mid \btheta)}{p(\bx)} = \log r(\bx \mid \btheta)}$ given unlimited model flexibility and data

\paragraph{\NREB}
\citet{Durkan2020} estimate $r(\bx \mid \btheta)$ by training a classifier that selects from among $K$ parameters $\bTheta \coloneqq (\btheta_1, \ldots, \btheta_K)$ which could have generated $\bx$. In contrast with \NREA, one of these parameters $\btheta_k$ is \emph{always} drawn jointly with $\bx$. Let $y$ be a random variable which indicates which one of $K$ parameters simulated $\bx$. The marginal probability $p(y=k) \coloneqq 1/K$ is uniform. That means

\begin{equation}
	p_{\NREB}(\bTheta, \bx \mid y = k) \coloneqq p(\btheta_1) \cdots p(\btheta_K) p(\bx \mid \btheta_k)
\end{equation}

defines our conditional probability for parameters and data. Bayes' rule reveals a conditional distribution over $y$, dropping the \NREB subscript, therefore

\begin{multline}
		p(y=k \mid \bTheta, \bx)
		= \frac{ p(\bTheta, \bx \mid y=k) p(y=k) }{ p(\bTheta, \bx) }
		= \frac{ p(\bTheta, \bx \mid y=k) p(y=k) }{ \sum_i p(\bTheta, \bx \mid y=i) p(y=i) }  \\
		= \frac{ p(\btheta_1) \cdots p(\btheta_k, \bx) \cdots p(\btheta_K) }{ \sum_i p(\btheta_1) \cdots p(\btheta_i,\bx) \cdots p(\btheta_K) } 
		= \frac{ p(\btheta_k \mid \bx)/p(\btheta_k) }{ \sum_i p(\btheta_i \mid \bx)/p(\btheta_i) } 
		= \frac{ r(\bx \mid \btheta_k) }{ \sum_i r(\bx \mid \btheta_i) }.
\end{multline}

\NREB estimates $\log \rhat(\bx \mid \btheta)$ with a neural network $g_{\bw}$. Training is done by minimizing the 
cross entropy of $q_{\bw}(y=k \mid \bTheta, \bx) \coloneqq \exp \circ g_{\bw}(\btheta_k, \bx) / \sum_i \exp \circ g_{\bw}(\btheta_i, \bx)$ relative to $p(y, \bx, \btheta)$;

\begin{equation}
    \label{eqn:nreb-loss-appndx}
    \bw = \argmin_{\bw}
        \mathbb{E}_{p(y, \bTheta, \bx)} \left[ - q_{\bw}(y \mid \bTheta, \bx) \right]
    	\approx \argmin_{\bw} \left[ - \frac{1}{B} \sum_{b'=1}^{B} \log \frac{ \exp \circ g_{\bw}(\btheta_{k}^{(b')}, \bx^{(b')}) }{ \sum_i \exp \circ g_{\bw}(\btheta_i^{(b')}, \bx^{(b')}) } \right]
\end{equation}

where $\btheta_{1}^{(b')}, \ldots, \btheta_{K}^{(b')} \sim p(\btheta)$, $k^{(b')} \sim p(y)$, and $\bx^{(b')} \sim p(\bx \mid \btheta_k^{(b')})$ over $B$ samples. In our parameterization and given unlimited flexibility and data, ${g_{\bw}(\btheta, \bx) = \log \frac{ p(\btheta \mid \bx) }{p(\btheta)} + c_{\bw}(\bx)}$ at convergence. The extra term enters because the optimal classifier for \eqref{eqn:nreb-loss-appndx} need not be normalized.

\section{Theoretical Arguments}
\label{apndx:proof}

We present first the proof of convergence for \NREC. Afterwards, we discuss the properties of estimated importance weights in \NREB.

\subsection{Proof of convergence of \NREC}
\begin{Lem}
\label{lem:optimal=log-ratio}
Consider for $k=0,\dots,K$ the following probability distributions for $\bz$:
\begin{align}
    p(\bz\mid y=k).
\end{align}
and $p(y) >0$ a probability distribution for $y$. Put $p_k:=p(y=k)$ for $k=1,\dots,K$.
For functions $f_k:\, \mathcal{Z} \to \mathbb{R}$, $k=1,\dots,K$,  let:
\begin{align}
        q(y = k \mid f, \bz) :=
    \begin{cases}
    	\frac{1}{1 +  \sum_{j=1}^{K} \frac{p_j}{p_0} \exp(f_j(\bz))}, & k = 0, \\
    	\frac{\frac{p_k}{p_0} \exp(f_k(\bz))}{1 +  \sum_{j=1}^{K} \frac{p_j}{p_0} \exp(f_j(\bz))}, & k = 1, \dots, K.
    \end{cases}
\end{align}
Note that $q(y = k \mid f, \bz) > 0$ for all $k=0,\dots,K$ and that $\sum_{k=0}^K q(y = k \mid f, \bz) =1$ for every $K$-tuple $f=(f_k)_{k=1,\dots,K}$ and $\bz \in \mathcal{Z}$. 
Consider a minimizer:
\begin{align}
f^* &\in \argmin_{f} \mathbb{E}_{p(\bz\mid y) p( y) } \left[ - \log q( y \mid f,\bz) \right].
\end{align}
Then we have for $p(\bz)$-almost-all $\bz$ and all $k=1,\dots,K$:
\begin{align}
   f^*_k(\bz) & = \log \frac{p(\bz| y =k)}{p(\bz| y =0)}.
\end{align}
\begin{proof}
We have:
\begin{align}
f^* &\in \argmin_{f} \mathbb{E}_{p(\bz\mid y ) p( y ) } \left[ - \log q( y  \mid f,\bz) \right] \\
&=\argmin_{f} \mathbb{E}_{p(\bz, y )} \left[ \log \frac{p( y \mid\bz)}{q( y  \mid f,\bz)} \right] \\
&=\argmin_{f} \mathbb{E}_{p(\bz)} \left[ \KLD(p( y \mid\bz)\| q( y  \mid f,\bz)) \right],
\end{align}
which is minimized, when $\KLD=0$, thus:
\begin{align}
    0 &= \KLD(p( y \mid\bz)\| q( y  \mid f^*,\bz)),
\end{align}
which implies that for $p(\bz)$-almost-all $\bz$:
\begin{align}
    q( y  \mid f^*,\bz) & = p( y \mid\bz) = \frac{p(\bz| y )}{p(\bz)}p( y ).
\end{align}
So we get with the definition of $q( y  \mid f^*,\bz)$:
\begin{align}
    \frac{p(\bz| y =0) }{p(\bz)} p_0 &= \frac{1}{1 +  \sum_{j=1}^{K} \frac{p_j}{p_0} \exp(f^*_j(\bz))}, & k = 0, \\
    \frac{p(\bz| y =k) }{p(\bz)}p_k &= \frac{\frac{p_k}{p_0} \exp(f^*_k(\bz))}{1 +  \sum_{j=1}^{K} \frac{p_j}{p_0} \exp(f^*_j(\bz))}, & k = 1, \dots, K.
\end{align}
Dividing the latter by the former gives for $k=1,\dots,K$:
\begin{align}
    \frac{p(\bz| y =k)}{p(\bz| y =0)} \frac{p_k}{p_0} &=\frac{p_k}{p_0} \exp(f^*_k(\bz)),
\end{align}
implying for $k=1,\dots,K$ and $p(\bz)$-almost-all $\bz$:
\begin{align}
   f_k^*(\bz) & = \log \frac{p(\bz| y =k)}{p(\bz| y =0)}.
\end{align}
This shows the claim.
\end{proof}
\end{Lem}

We used the symbol \KLD to imply the Kullback-Leibler Divergence. The proof in Lemma~\ref{lem:optimal=log-ratio} is slightly more general than necessary for our typical case. All our functions $f_k$ are typically the same, namely the evaluation of a neural network with weights $\bw$. Rather than searching for the function $f$ which minimizes the objective, we search for the weights, but these are equivalent. Finally to make everything fit, we set $\bz \coloneqq (\btheta, \bx)$. 

\subsection{Properties of the importance sampling diagnostic on biased ratio estimates}

In Section~\ref{sec:diagnostic} we discuss the importance sampling diagnostic, in which the estimated ratio is tested by comparing weighted samples from $\bx \sim p(\bx)$ to unweighted samples from $\bx \sim p(\bx \mid \btheta)$. The estimated ratio from \NREC is merely $\exp(h_{\bw}(\btheta, \bx))$ and is therefore not restricted to have the properties of a ratio of probability distributions, except at optimum with unlimited flexibility and data. The importance sampling diagnostic is designed to test whether the estimated ratio is close to having these properties for a fixed $\btheta$. One way to improve ratios estimated by \NREC is to compute the normalizing constant $Z_{\bw}(\bx)$, which should be close to one, and replace the ratio with this ``normalized'' version.

The unrestricted nature of the \NREB-specific $c_{\bw}(\bx)$ bias means that the normalization constant $Z_{\bw}(\bx)$ does not have to be close to one. Further, we show that the unrestricted bias means that the normalizing constant for two \NREB ratio estimators, which are equivalent in terms of their loss function \eqref{eqn:nreb-loss-appndx}, is not unique, i.e., the estimate is ill-posed. This property means that normalizing \NREB will not, in general, produce an estimator which passes the diagnostic.

Consider two ratio estimates with the relationship $\rhat_1(\bx \mid \btheta) = \rhat_2(\bx \mid \btheta) / C(\bx)$ where $C$ is an positive function of $\bx$ resulting from the aggregation of the exponentiated bias. Given $N$ samples $\bx_n \sim p(\bx)$ with weights $w_1(\bx_n) = \rhat_1(\bx_n \mid \btheta)$ and $w_2(\bx_n) = \rhat_2(\bx_n \mid \btheta)$, we can compute the importance-normalized weights by 

\begin{align}
    \bar{w}_{1}(\bx_n) &= \frac{w_1(\bx_n)}{\sum_{i=1}^N w_1(\bx_i)}, 
    &
    \bar{w}_{2}(\bx_n) &= \frac{w_2(\bx_n)}{\sum_{i=1}^N w_2(\bx_i)}.
\end{align}

However, if we substitute this constant back into our expression, we find that the weights do not agree

\begin{align}
    \bar{w}_{1}(\bx_n) 
    = \frac{w_2(\bx_n) / C(\bx_n)}{\sum_{i=1}^N w_2(\bx_i) / C(\bx_i)}
    &\neq \bar{w}_{2}(\bx_n).
\end{align}

Therefore, normalization does not ``protect'' against scaling bias introduced by a function in $\bx$. \NREB does not penalize functional biases like $C(\bx)$, so even ratios considered optimal by \NREB can easily fail the diagnostic. Meanwhile, \NREC encourages terms like $C(\bx)$ towards one. That has the effect of making the $C(\bx) \approx 1$ drop out of the normalized importance weights, thus making performance on the diagnostic indicative of a better ratio estimate, given enough classifier flexibility.

\section{Experimental Details}
\label{apndx:experimental-details}

\paragraph{Computational costs}
Our experiments were performed on a cluster of Nvidia Titan V graphics processing units. The primary expense was the hyperparameter search within Sections~\ref{sec:unlimited-joint}, \ref{sec:unlimited-prior}, and the first part of Section~\ref{sec:benchmark}. The run of those experiments took about 50,000 gpu-minutes total considering all of our current data. Since there were several iterations, we multiply this by three to estimate total compute. The next expense was the computation of \texttt{sbibm} in Section~\ref{sec:benchmark} that took about 24,000 gpu-minutes. Together these equal about 2880 gpu-hours. An estimate for the total carbon contributions corresponds to 311.04 kgCO$_2$. Luckily, our clusters run completely on wind power offsetting the contribution. Estimations were conducted using the \href{https://mlco2.github.io/impact#compute}{MachineLearning Impact calculator} presented in \cite{lacoste2019quantifying}.

\paragraph{Architecture and training}
The centerpiece of our method is the neural network $h_{\bw}$ which is trained as a classifier. The hyperparameter choices here were fairly constant throughout the experiments. Any hyperparameters for training or architecture which were consistent across experiments are listed in Table~\ref{table:architecture}. Hidden features and number of \textsc{resnet} blocks depend on the architecture. Large NN uses three resnet blocks with 128 features, while Small NN uses two resnet blocks and 50 features.

\begin{table}[htb]
	\caption{Architecture}
	\label{table:architecture}
	\centering
    \begin{tabular}{ll}
        \toprule
        Hyperparameter & Value \\
        \midrule
        Activation Function & \textsc{relu} \\
        \textsc{amsgrad} & No \\
        Architecture & \textsc{resnet} \\
        Batch normalization & Yes \\
        Batch size & 1024 \\
        Dropout & No \\
        Max epochs & 1000 \\
        Learning rate & 0.0005 \\
        Optimizer & \textsc{adam} \\
        Weight Decay & 0.0 \\
        Standard-score Observations & Using first batch \\
        Standard-score Parameters & Using first batch \\
        \bottomrule
    \end{tabular}
\end{table}

\subsection{Hyperparameter search measured with C2ST}
\begin{figure}[htb]
    \centering
    \includegraphics[width=1.0\textwidth]{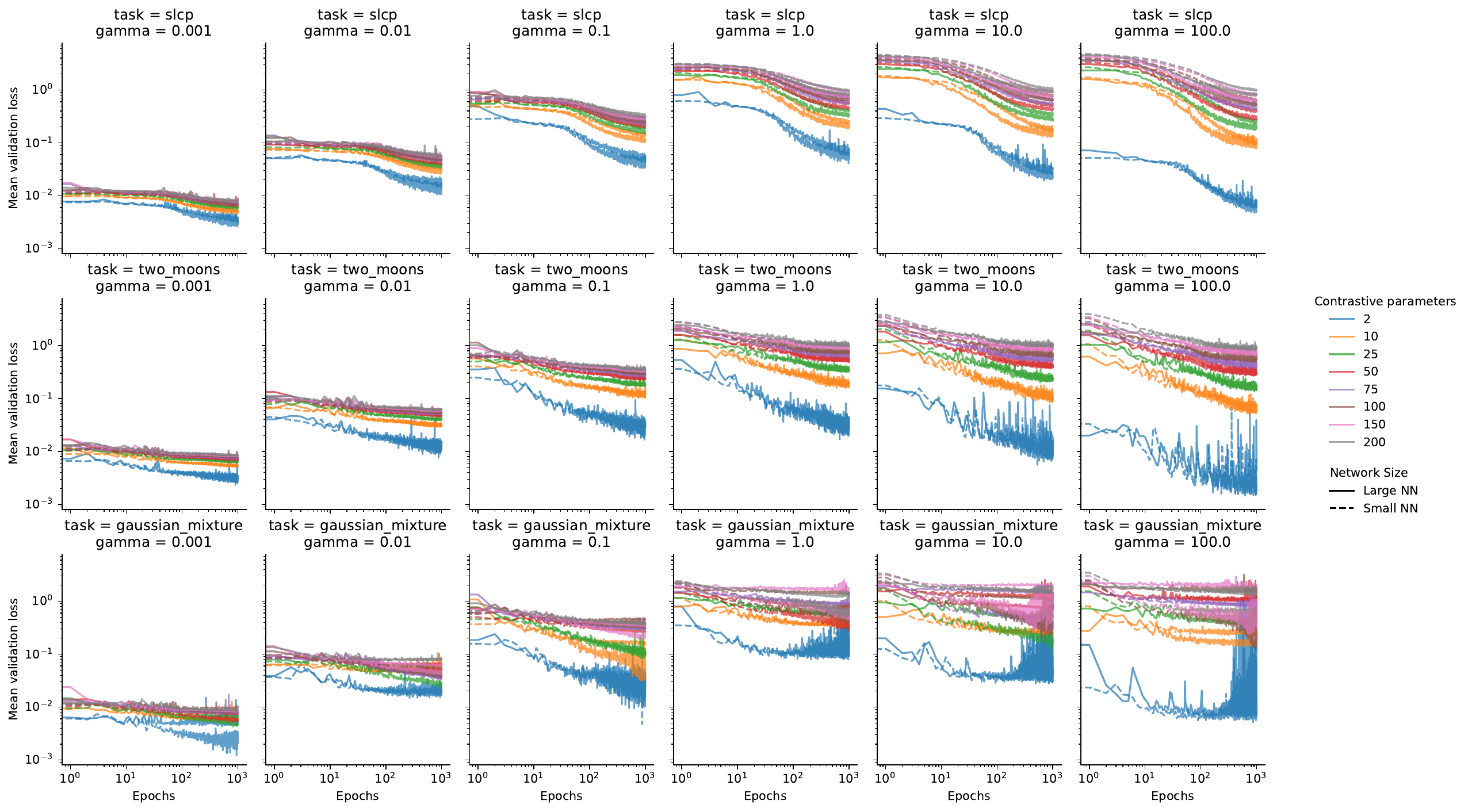}
    \caption{
    The validation loss from \NREC is reported versus epochs on various tasks, $\gamma$, $K$, and architectures trained using unlimited draws from the joint distribution. The rows correspond to different tasks, columns to different $\gamma$, colors to different $K$, and dashed or solid lines to Small and Large NN respectively. These plots correspond with the technique discussed in Section~\ref{sec:unlimited-joint}.
    }
    \label{fig:grid-validation-joint}
\end{figure}

\begin{figure}[htb]
    \centering
    \includegraphics[width=1.0\textwidth]{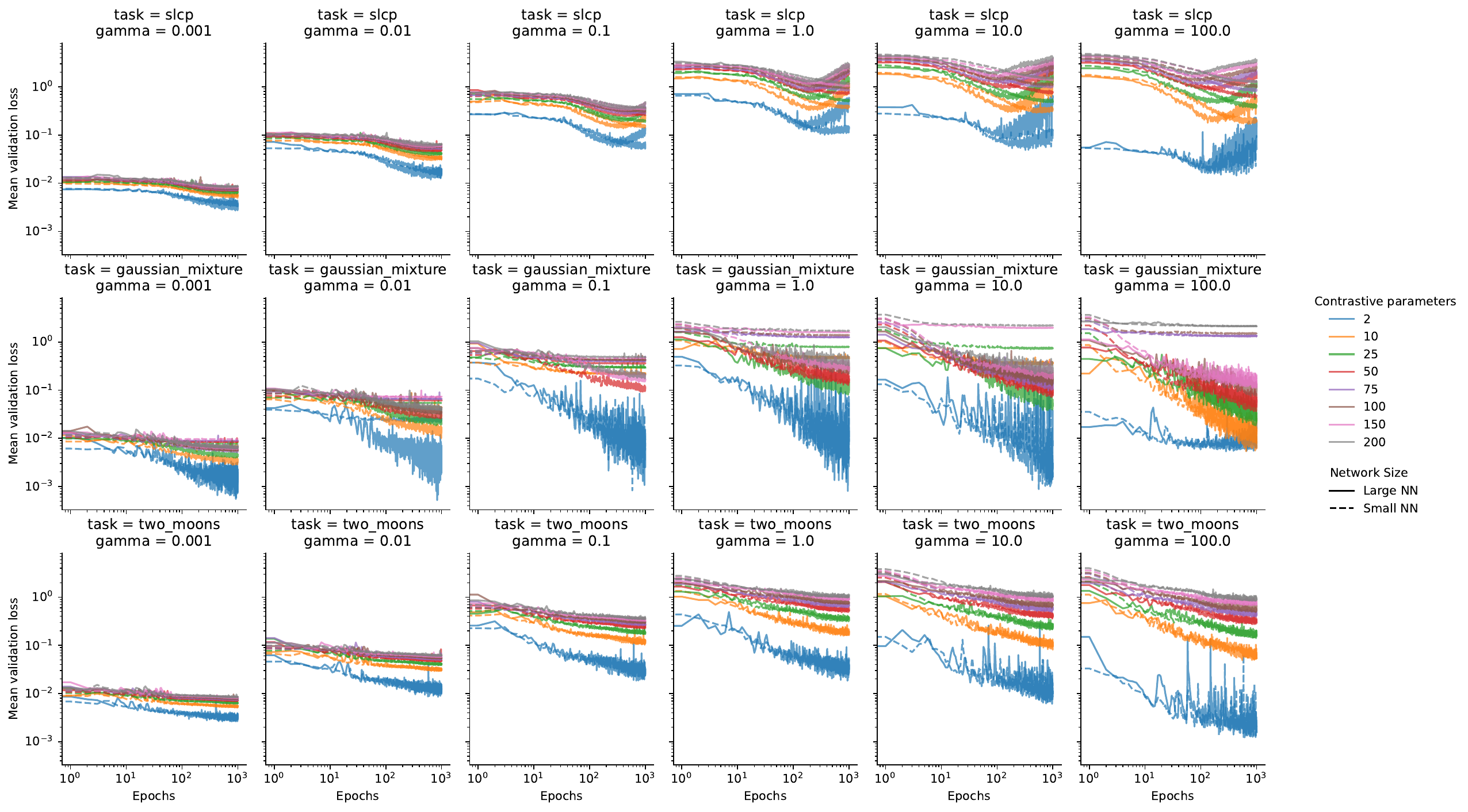}
    \caption{
    The validation loss from \NREC is reported versus epochs on various tasks, $\gamma$, $K$, and architectures trained using a simulation budget of 22,528 but unlimited draws from the prior during training. The rows correspond to different tasks, columns to different $\gamma$, colors to different $K$, and dashed or solid lines to Small and Large NN respectively. These plots correspond with the technique discussed in Section~\ref{sec:unlimited-prior}.
    }
    \label{fig:grid-validation-prior}
\end{figure}

\begin{figure}[htb]
    \centering
    \includegraphics[width=1.0\textwidth]{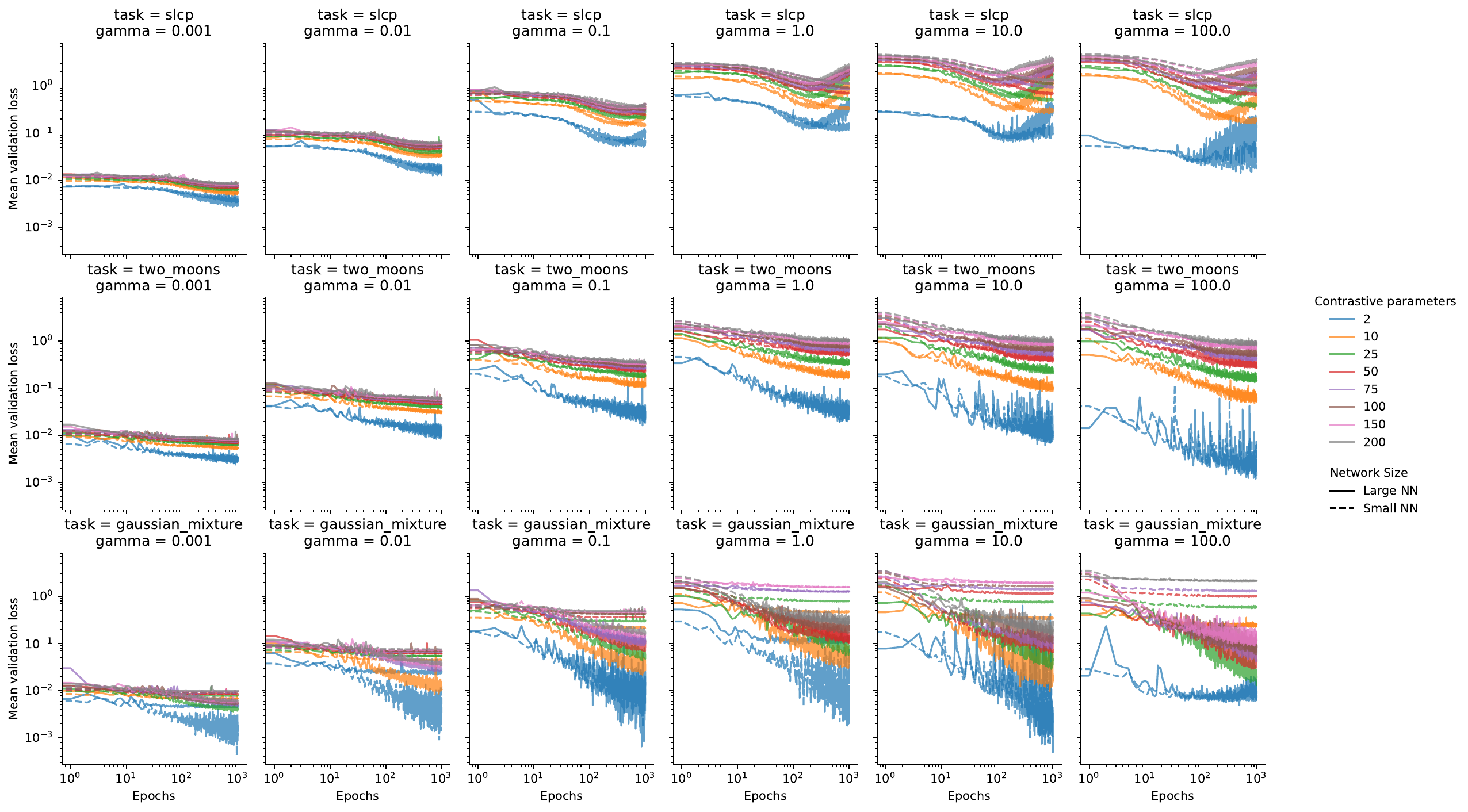}
    \caption{
    The validation loss from \NREC is reported versus epochs on various tasks, $\gamma$, $K$, and architectures trained using a fixed simulation budget of 22,528. The rows correspond to different tasks, columns to different $\gamma$, colors to different $K$, and dashed or solid lines to Small and Large NN respectively. These plots correspond with the technique discussed in Section~\ref{sec:benchmark}.
    }
    \label{fig:grid-validation-bench}
\end{figure}

In this section, we trained many neural networks with different hyperparameter settings on three different tasks from \citet{sbibm}. We trained both architectures, Large and Small NN, on a grid of $\gamma$ and $K$ values with \NREB and \NREC. No matter whether we were training with unlimited draws from the joint, prior, or fixed data we designed an epoch such that it has 20 training mini-batches and 2 validation mini-batches. For fixed initial data and prior this corresponds to a simulation budget of 22,528. The mean validation losses per-epoch for these networks is visible in Figures~\ref{fig:grid-validation-joint}, \ref{fig:grid-validation-prior}, and \ref{fig:grid-validation-bench}. 

\begin{figure}[htb]
    \centering
    \includegraphics[width=1.0\textwidth]{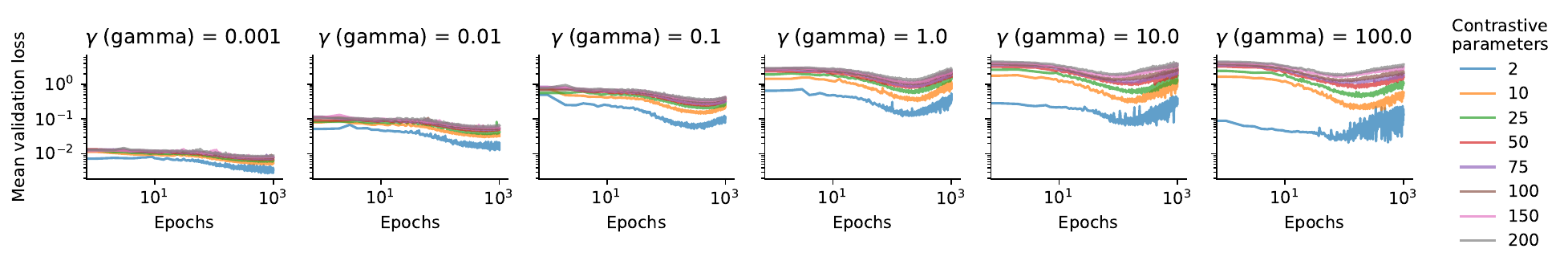}
    \caption{
    The validation loss from \NREC is reported versus epochs on SLCP with the Large NN where colors indicate different contrastive parameter counts. The plot shows the convergence rates of each model. Since validation loss is a function of $K$ and $\gamma$, the relative performance of models is \emph{not} comparable. 
    A grid search of $K$ and $\gamma$ indicates that increasing $K$ leads to earlier convergence at fixed $\gamma$. With $K$ fixed,  $\gamma < 1$ has a negative effect on convergence rate and $\gamma > 1$ is ambiguous.
    }
    \label{fig:slcp-validation-bench}
\end{figure}

\begin{figure}[htb]
    \centering
    \includegraphics[width=1.0\textwidth]{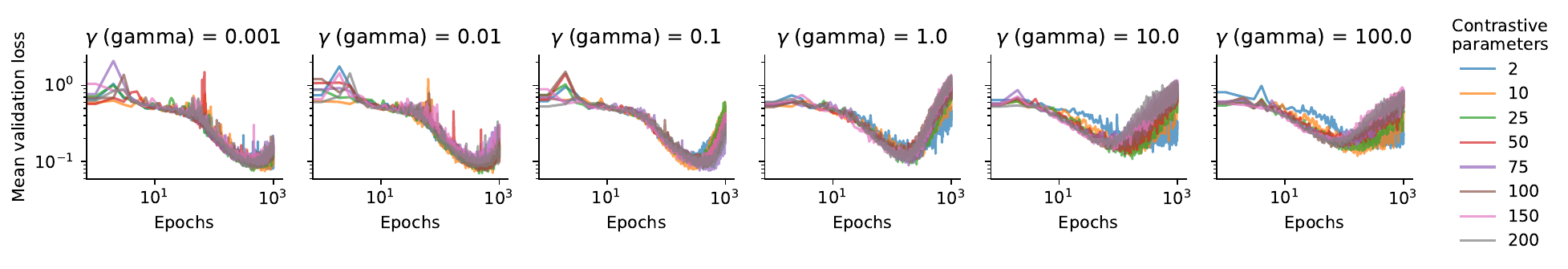}
    \caption{
    The validation loss from \NREC is reported versus epochs on SLCP with the Large NN where colors indicate different contrastive parameter counts, i.e, $K + 1$.
    A fixed validation loss was used, namely $\ell_{1,1}$. Although the validation loss is now comparable so that we can see the performance of the different classifiers on the same task, some of the convergence rate trends disagreed with SLCP in Figure~\ref{fig:slcp-validation-bench}.
    }
    \label{fig:slcp-validation-sameval}
\end{figure}

For one specific problem, we also computed validation loss at a fixed $\gamma$ and $K$ no matter the training regime in an effort to produce comparable validation losses, specifically $\gamma=1$ and $K=1$ aka the \NREA regime. This training setting reverses some of the trends we've seen when validated on the same loss as training data. We did not apply it further since this could bias results. See Figure~\ref{fig:slcp-validation-sameval}. To compare, we also plot the validation loss on the sample problem but using the $\gamma$ and $K$ each model was trained with. See Figure~\ref{fig:slcp-validation-bench}, note that the bias depending on $\gamma$ and $K$ is clearly visible (just as in the other validation loss plots).

\begin{figure}[htb]
\centering
    \begin{subfigure}[b]{0.30\textwidth}
        \includegraphics[width=\textwidth]{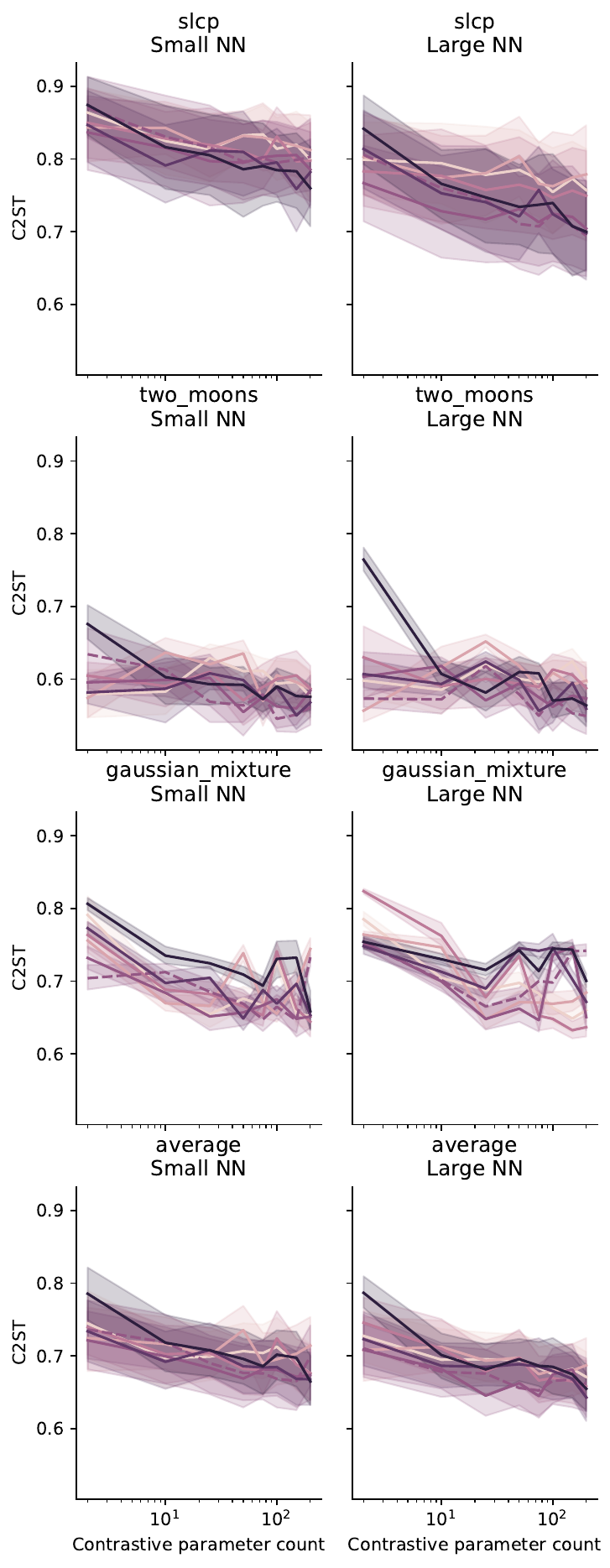}
        \caption{Draw $(\btheta, \bx)$.}
    \end{subfigure}
    \hfill
    \begin{subfigure}[b]{0.30\textwidth}
        \includegraphics[width=\textwidth]{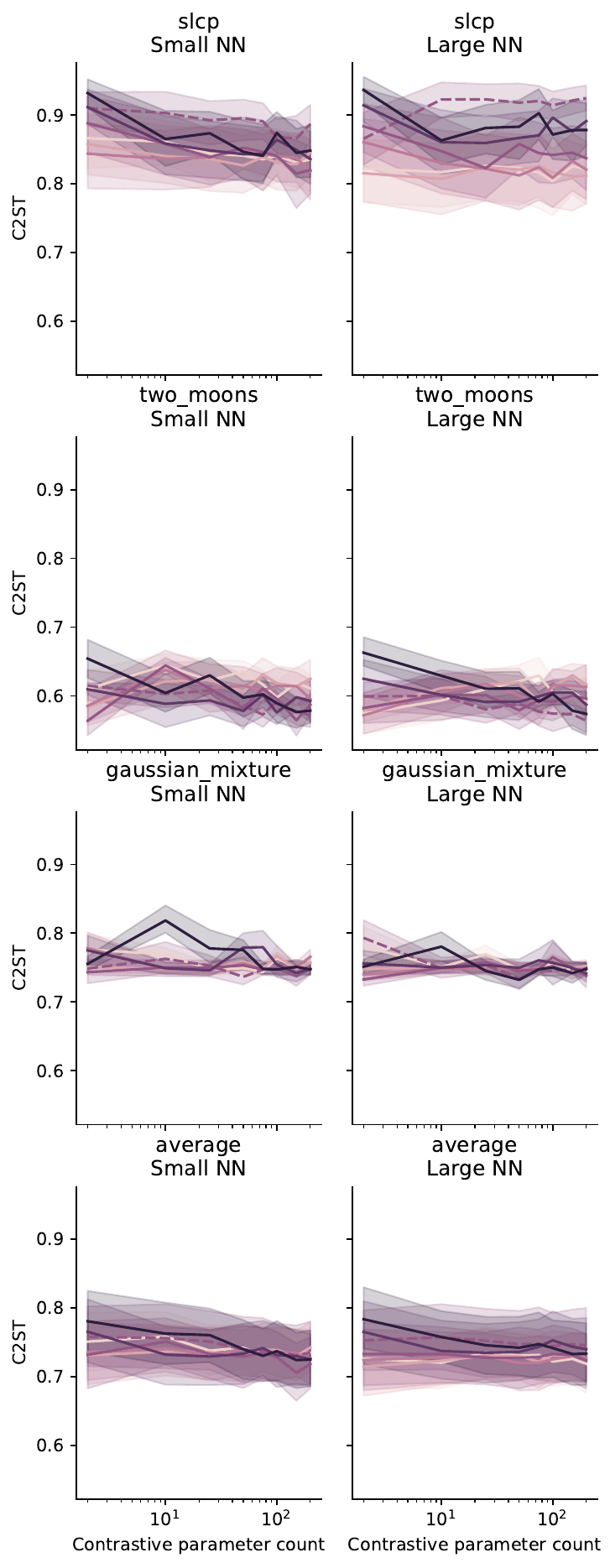}
        \caption{Draw $\btheta$.}
    \end{subfigure}
    \hfill
    \begin{subfigure}[b]{0.375\textwidth}
        \includegraphics[width=\textwidth]{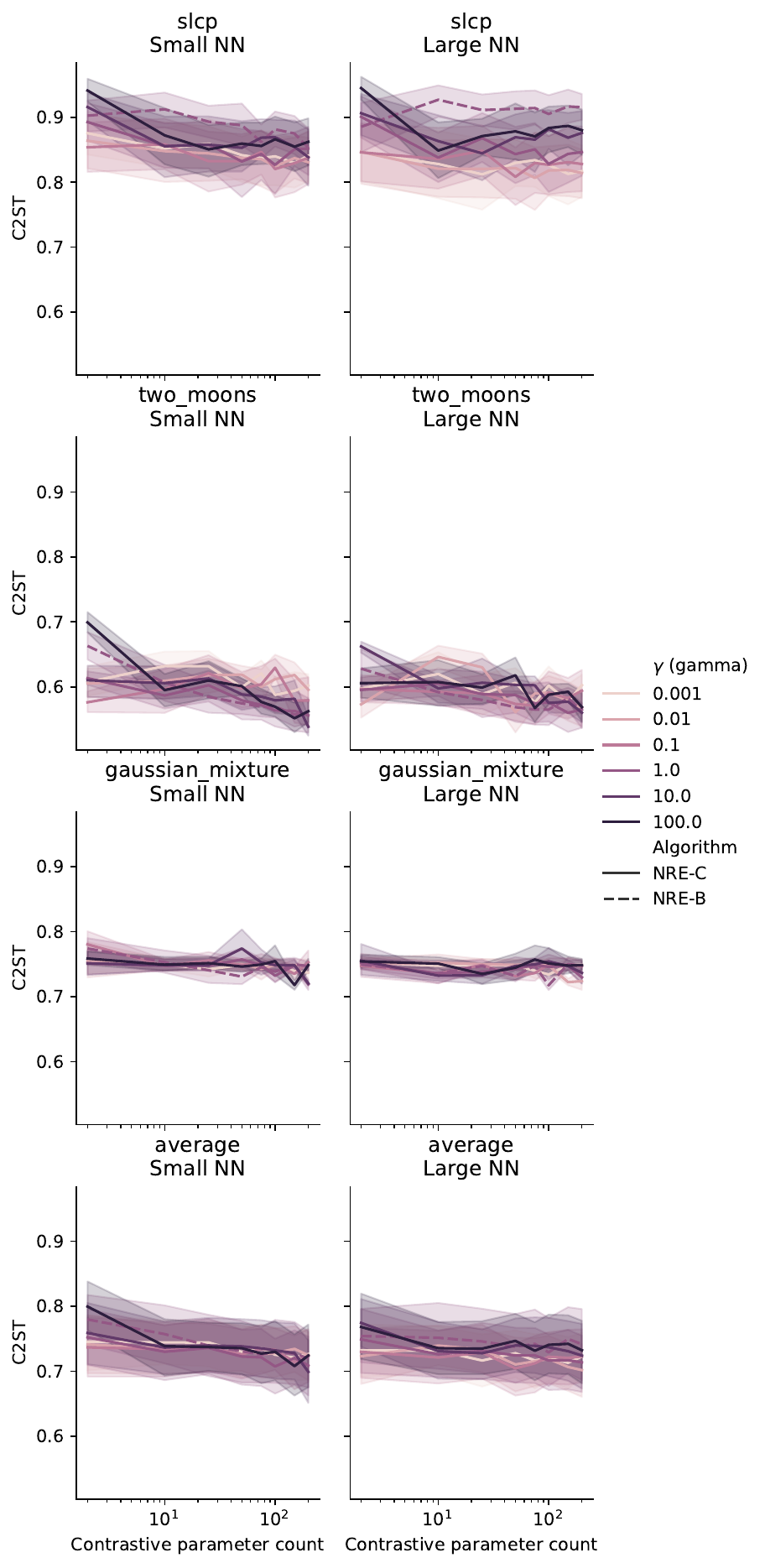}
        \caption{Fixed data.}
    \end{subfigure}
    \caption{%
        A measure of exactness comparing to the ground truth to the samples from a surrogate posterior, the C2ST, is plotted as a function of number of contrastive parameters. Each row corresponds to a different task: SLCP, Two Moons, Gaussian Mixture, and an average of the results across tasks. Both \NREC and \NREB are shown along with various $\gamma$ values, and architectures are shown. Recall that C2ST assigns 1.0 to inaccurate and 0.5 to accurate approximations. (a) Corresponds to Section~\ref{sec:unlimited-joint} where unlimited draws from the joint are allowed. (b) Corresponds to Section~\ref{sec:unlimited-prior} where unlimited draws from the prior are allowed but the $\bx$ data is fixed. (c) Corresponds to Section~\ref{sec:benchmark} where both the initial draws of $\btheta$ and $\bx$ are the only data available.
    }
    \label{fig:c2st-specific}
\end{figure}

Once the networks were trained, we drew samples from ten per-task posteriors based on ten predefined observations $\bx_i$ with $i \in 1, 2, \ldots, 10$. This leveraged the amortized property of \NREC. Samples were drawn on these problems using rejection sampling. Once the samples were drawn, they were compared to ground truth posterior samples from the benchmark with the C2ST. The detailed plot which shows per-task behavior is available in Figure~\ref{fig:c2st-specific}.

\clearpage

\subsection{Simulation-based inference benchmark}

We trained \NREC with $\gamma = 1.0$ and $K=100$ using Large NN for 1000 epochs on all of the tasks from the simulation-based inference benchmark \cite{sbibm}. In contrast with our previous hyperparameter search experiments, we used their Markov-chain Monte Carlo sampling scheme which applies slice sampling \cite{neal2003slice}. The details results are presented in Table~\ref{table:sbibm-details} and Table~\ref{table:sbibm-details-2}. There were three tasks which did not succeed with this sampling procedure, there we used rejection sampling. 

Our training slightly diverged from the benchmark. We enumerate the ways below.
\begin{itemize}
	\item Large NN is a bigger architecture than the one which produced their reported \NREB values. In fact, the \NREB version has the same settings as Small NN; however, we found that larger networks performed better in our search and chose the bigger one for that reason.
	\item In the benchmark, networks are trained with early stopping but we always trained for 1000 epochs. We selected the network which performed best on the validation loss, using the same values for $K$ and $\gamma$ in the validation loss as in the training.
	\item Even though \NREB is an amortized method, the benchmark trained a network for every observation. Instead, we leveraged the amortization properties of \NREC and trained a single network which drew samples from each approximate posterior given by each of the ten observations.
\end{itemize}

\paragraph{Task details}
We provide a short summary of all of the inference tasks in the \SBI benchmark by \citet{sbibm}.

\begin{itemize}[align=left]
	\item[\textbf{Bernoulli GLM}]  This task is a generalized linear model. The likelihood is Bernoulli distributed. The data is a 10-dimensional summary statistic from an 100-dimensional raw vector. The posterior is 10-dimensional and it only has one mode.
	\item[\textbf{Bernoulli GLM Raw}]  This is the same task as above, but instead the entire 100-dimensional observation is shown to the inference method rather than the summary statistic.
	\item[\textbf{Gaussian Linear}]  A simple task with a Gaussian distributed prior and a Gaussian likelihood over the mean. Both have a $\Sigma=0.1 \cdot \identity$ covariance matrix. The posterior is also Gaussian. It is performed in 10-dimensions for the observations and parameters.
	\item[\textbf{Gaussian Linear Uniform}]  This is the same as the task above, but instead the prior over the mean is a 10-dimensional uniform distribution from -1 to 1 in every dimension.
	\item[\textbf{Gaussian Mixture}]  This task occurs in the ABC literature often. Infer the common mean of a mixture of Gaussians where one has covariance matrix $\Sigma=1.0 \cdot \identity$ and the other $\Sigma=0.01 \cdot \identity$. It occurs in two dimensions.
	\item[\textbf{Lotka Volterra}]  This is an ecological predator-prey model where the simulations are generated from randomly drawn initial conditions by solving a parameterized differential equation. There are four parameters that control the coupling between the generation and destruction of both prey and predators. The priors are log normal. The data is a twenty dimensional summary statistic.
	\item[\textbf{SIR}]  An epidemiological model simulating the progress of an contagious disease outbreak through a population. Simulations are generated from randomly drawn initial conditions with a parameterized differential equation defining the dynamics. There are two parameters with a log normal prior. The data is a ten dimensional summary statistic.
	\item[\textbf{SLCP}]  A task which has a very simple non-spherical Gaussian likelihood, but a complex posterior over the five parameters which, via a non-linear function, define the mean and covariance of the likelihood. There are five parameters each with a uniform prior from -3 to 3. The data is four-dimensional but we take two samples from it. It was introduced in \cite{papamakarios2019sequential}.
	\item[\textbf{SLCP with Distractors}] This is the same task as above but instead the data is concatenated with 92 dimensions of Gaussian noise.
	\item[\textbf{Two Moons}]  This task exhibits a crescent shape posterior with bi-modality--two of the attributes often used to stump \MCMC samplers. Both the data and parameters are two dimensional. The prior is uniform from -1 to 1.
\end{itemize}

\begin{table}
\centering
\caption{Simulation-based inference benchmark results.}
\label{table:sbibm-details}
\begin{tabular}{llrrr}
\toprule
 &  & \multicolumn{3}{r}{C2ST} \\
 & Simulation budget & $10^3$ & $10^4$ & $10^5$ \\
Task & Algorithm &  &  &  \\
\midrule
\multirow[c]{9}{*}{Bernoulli GLM} & \NREC (ours) & 0.802 & 0.694 & 0.601 \\
 & \REJABC & 0.994 & 0.976 & 0.941 \\
 & \NLE & 0.740 & 0.605 & 0.545 \\
 & \NPE & 0.863 & 0.678 & 0.559 \\
 & \NRE (\NREB) & 0.899 & 0.812 & 0.751 \\
 & \SMCABC & 0.991 & 0.981 & 0.818 \\
 & \SNLE & 0.634 & 0.553 & 0.522 \\
 & \SNPE & 0.855 & 0.614 & 0.525 \\
 & \SNRE (\SNREB) & 0.718 & 0.584 & 0.529 \\
\cline{1-5}
\multirow[c]{9}{*}{Bernoulli GLM Raw} & \NREC (ours) & 0.928 & 0.775 & 0.609 \\
 & \REJABC & 0.995 & 0.984 & 0.966 \\
 & \NLE & 0.870 & 0.939 & 0.951 \\
 & \NPE & 0.900 & 0.765 & 0.607 \\
 & \NRE (\NREB) & 0.915 & 0.834 & 0.777 \\
 & \SMCABC & 0.990 & 0.959 & 0.943 \\
 & \SNLE & 0.990 & 0.973 & 0.987 \\
 & \SNPE & 0.906 & 0.658 & 0.607 \\
 & \SNRE (\SNREB) & 0.880 & 0.675 & 0.552 \\
\cline{1-5}
\multirow[c]{9}{*}{Gaussian Linear} & \NREC (ours) & 0.655 & 0.556 & 0.519 \\
 & \REJABC & 0.913 & 0.858 & 0.802 \\
 & \NLE & 0.650 & 0.555 & 0.515 \\
 & \NPE & 0.694 & 0.552 & 0.506 \\
 & \NRE (\NREB) & 0.672 & 0.560 & 0.536 \\
 & \SMCABC & 0.922 & 0.829 & 0.726 \\
 & \SNLE & 0.628 & 0.548 & 0.519 \\
 & \SNPE & 0.652 & 0.544 & 0.507 \\
 & \SNRE (\SNREB) & 0.670 & 0.536 & 0.515 \\
\cline{1-5}
\multirow[c]{9}{*}{Gaussian Linear Uniform} & \NREC (ours) & 0.746 & 0.660 & 0.550 \\
 & \REJABC & 0.977 & 0.948 & 0.909 \\
 & \NLE & 0.723 & 0.548 & 0.506 \\
 & \NPE & 0.696 & 0.553 & 0.509 \\
 & \NRE (\NREB) & 0.788 & 0.706 & 0.631 \\
 & \SMCABC & 0.968 & 0.928 & 0.794 \\
 & \SNLE & 0.657 & 0.552 & 0.509 \\
 & \SNPE & 0.631 & 0.527 & 0.507 \\
 & \SNRE (\SNREB) & 0.681 & 0.606 & 0.536 \\
\cline{1-5}
\multirow[c]{9}{*}{Gaussian Mixture} & \NREC (ours) & 0.757 & 0.750 & 0.724 \\
 & \REJABC & 0.883 & 0.789 & 0.772 \\
 & \NLE & 0.812 & 0.731 & 0.757 \\
 & \NPE & 0.731 & 0.661 & 0.555 \\
 & \NRE (\NREB) & 0.784 & 0.752 & 0.734 \\
 & \SMCABC & 0.799 & 0.746 & 0.664 \\
 & \SNLE & 0.701 & 0.702 & 0.624 \\
 & \SNPE & 0.697 & 0.583 & 0.533 \\
 & \SNRE (\SNREB) & 0.723 & 0.662 & 0.542 \\
\cline{1-5}
\bottomrule
\end{tabular}
\end{table}

\begin{table}
\centering
\caption{Simulation-based inference benchmark results continued.}
\label{table:sbibm-details-2}
\begin{tabular}{llrrr}
\toprule
 &  & \multicolumn{3}{r}{C2ST} \\
 & Simulation budget & $10^3$ & $10^4$ & $10^5$ \\
Task & Algorithm &  &  &  \\
\midrule
\multirow[c]{9}{*}{Lotka-Volterra} & \NREC (ours) & 0.999 & 0.981 & 0.974 \\
 & \REJABC & 1.000 & 1.000 & 0.998 \\
 & \NLE & 0.994 & 0.956 & 0.952 \\
 & \NPE & 0.999 & 0.997 & 0.981 \\
 & \NRE (\NREB) & 1.000 & 0.998 & 0.996 \\
 & \SMCABC & 1.000 & 0.996 & 0.995 \\
 & \SNLE & 0.909 & 0.738 & 0.695 \\
 & \SNPE & 0.990 & 0.953 & 0.928 \\
 & \SNRE (\SNREB) & 0.971 & 0.848 & 0.831 \\
\cline{1-5}
\multirow[c]{9}{*}{SIR} & \NREC (ours) & 0.761 & 0.625 & 0.995 \\
 & \REJABC & 0.964 & 0.838 & 0.713 \\
 & \NLE & 0.761 & 0.748 & 0.730 \\
 & \NPE & 0.815 & 0.680 & 0.585 \\
 & \NRE (\NREB) & 0.841 & 0.770 & 0.690 \\
 & \SMCABC & 0.921 & 0.626 & 0.613 \\
 & \SNLE & 0.745 & 0.745 & 0.650 \\
 & \SNPE & 0.638 & 0.561 & 0.575 \\
 & \SNRE (\SNREB) & 0.637 & 0.646 & 0.547 \\
\cline{1-5}
\multirow[c]{9}{*}{SLCP} & \NREC (ours) & 0.974 & 0.892 & 0.794 \\
 & \REJABC & 0.982 & 0.973 & 0.961 \\
 & \NLE & 0.946 & 0.771 & 0.699 \\
 & \NPE & 0.975 & 0.901 & 0.831 \\
 & \NRE (\NREB) & 0.972 & 0.947 & 0.919 \\
 & \SMCABC & 0.982 & 0.969 & 0.963 \\
 & \SNLE & 0.921 & 0.713 & 0.578 \\
 & \SNPE & 0.965 & 0.845 & 0.666 \\
 & \SNRE (\SNREB) & 0.968 & 0.917 & 0.721 \\
\cline{1-5}
\multirow[c]{9}{*}{SLCP Distractors} & \NREC (ours) & 0.983 & 0.953 & 0.766 \\
 & \REJABC & 0.988 & 0.987 & 0.987 \\
 & \NLE & 0.987 & 0.961 & 0.905 \\
 & \NPE & 0.982 & 0.970 & 0.863 \\
 & \NRE (\NREB) & 0.980 & 0.968 & 0.953 \\
 & \SMCABC & 0.986 & 0.987 & 0.985 \\
 & \SNLE & 0.992 & 0.949 & 0.883 \\
 & \SNPE & 0.978 & 0.931 & 0.778 \\
 & \SNRE (\SNREB) & 0.981 & 0.974 & 0.766 \\
\cline{1-5}
\multirow[c]{9}{*}{Two Moons} & \NREC (ours) & 0.680 & 0.578 & 0.520 \\
 & \REJABC & 0.960 & 0.847 & 0.664 \\
 & \NLE & 0.773 & 0.713 & 0.668 \\
 & \NPE & 0.725 & 0.606 & 0.542 \\
 & \NRE (\NREB) & 0.822 & 0.761 & 0.629 \\
 & \SMCABC & 0.922 & 0.707 & 0.663 \\
 & \SNLE & 0.657 & 0.571 & 0.582 \\
 & \SNPE & 0.643 & 0.554 & 0.530 \\
 & \SNRE (\SNREB) & 0.651 & 0.582 & 0.563 \\
\cline{1-5}
\bottomrule
\end{tabular}
\end{table}

\clearpage

\section{Mutual Information}
\label{apndx:mutual-information}

Estimating the mutual information is closely related to estimating the likelihood-to-evidence ratio. We reference various bounds on the mutual information that and we can estimate using ratios learned by \NREC. These bounds obey the variational principle and might be a practical candidate for validating the performance of \SBI methods for scientific purposes, i.e., when the ground truth posterior is intractable. The approximation of the mutual information could synergize with other diagnostics like empirical, expected coverage testing and the importance sampling diagnostic. See section~\ref{sec:diagnostic}.

\paragraph{Mutual information and expected Kullback-Leibler divergence}
If $p(\btheta \mid \bx)$ denotes the true posterior and $p_{\bw}(\btheta \mid \bx)$ an approximate posterior then the quality of that approximation can be measured via the (forward) Kullback-Leibler divergence:
\begin{align}
    \KLD(p(\btheta \mid \bx) \Mid p_{\bw}(\btheta \mid \bx)).
\end{align}
However, in the \SBI-setting we have only access to the likelihood $p(\bx\mid \btheta)$ via samples. Since we have also access to the prior $p(\btheta)$ (analytically and) via samples we can sample from the joint $p(\btheta,\bx)$.
So using the \emph{expected Kullback-Leibler divergence} is more tractable in the \SBI setting to measure the discrepancy between true and approximated prior:
\begin{align}
   \mathbb{E}_{p(\bx)} \left[ \KLD(p(\btheta \mid \bx) \Mid p_{\bw}(\btheta \mid \bx))\right].
\end{align}
In all considered cases in this paper the approximate posterior $p_{\bw}(\btheta \mid \bx)$ is given by:
\begin{align}
    p_{\bw}(\btheta \mid \bx) & = \frac{\rhat_{\bw}(\bx \mid \btheta)}{Z_{\bw}(\bx)} p(\btheta), & Z_{\bw}(\bx) &:= \int \rhat_{\bw}(\bx \mid \btheta) p(\btheta) \,d\btheta,
\end{align}
where $\rhat_{\bw}(\bx \mid \btheta)$ comes from a (trained) neural network and $Z_{\bw}(\bx)$ denotes the normalization constant. 
With the above notations we get for the expected Kullback-Leibler divergence the expression:
\begin{align}
   \mathbb{E}_{p(\bx)} \left[ \KLD(p(\btheta \mid \bx) \Mid p_{\bw}(\btheta \mid \bx))\right] 
    &= \mathbb{E}_{p(\btheta, \bx)} \left[ 
		\log \frac{p(\btheta \mid \bx)}{p_{\bw}(\btheta \mid \bx)} \right]\\
    &= \mathbb{E}_{p(\btheta, \bx)} \left[ 
		\log \frac{p(\btheta,\bx)}{p(\btheta)p(\bx)} \frac{p(\btheta)}{p_{\bw}(\btheta \mid \bx)} \right] \\
	&= I(\btheta; \bx) - \mathbb{E}_{p(\btheta, \bx)} \left[ 
		\log \frac{\rhat_{\bw}(\bx \mid \btheta)}{Z_{\bw}(\bx)} \right] \\
	&= I(\btheta; \bx) - \mathbb{E}_{p(\btheta, \bx)} \left[ 
		\log \rhat_{\bw}(\bx \mid \btheta) \right] + \mathbb{E}_{p(\bx)} \left[ 
		\log Z_{\bw}(\bx) \right],
\end{align}
where $I(\btheta; \bx)$ is the \emph{mutual information} w.r.t.\ $p(\btheta,\bx)$.
Since we aim at minimizing the expected Kullback-Leibler divergence we implicitly aim to maximize our mutual information approximation:
\begin{align}
    I_{\bw}^{(0)}(\btheta; \bx) 
    &:= \mathbb{E}_{p(\btheta, \bx)} \left[ \log \rhat_{\bw}(\bx \mid \btheta) \right] - \mathbb{E}_{p(\bx)} \left[ \log Z_{\bw}(\bx) \right] \\
	&= \mathbb{E}_{p(\btheta, \bx)} \left[ \log \rhat_{\bw}(\bx \mid \btheta) \right] - \mathbb{E}_{p(\bx)} \left[ \log \mathbb{E}_{p(\btheta)} [\rhat_{\bw}(\bx \mid \btheta)] \right]\\
	&= \mathbb{E}_{p(\btheta, \bx)} \left[ h_{\bw}(\btheta,\bx) \right] - \mathbb{E}_{p(\bx)} \left[ \log \mathbb{E}_{p(\btheta)} [\exp(h_{\bw}(\btheta,\bx))] \right],
\end{align}
which can be estimated (with a bias due to the $\log$ in the expectation) via Monte-Carlo by sampling i.i.d.\ $\btheta_n,\btheta_{n,m} \sim p(\btheta)$, $\bx_n \sim p(\bx\mid \btheta_n)$, $n=1,\dots,N$, $m=1,\dots,M$ and then compute:
\begin{align}
    \hat{I}_{\bw}^{(0)}(\btheta; \bx) 
    &:= \frac{1}{N} \sum_{n=1}^N \log \rhat_{\bw}(\bx_n \mid \btheta_n) 
    - \frac{1}{N} \sum_{n=1}^N \log \left( \frac{1}{M} \sum_{m=1}^M \rhat_{\bw}(\bx_n \mid \btheta_{n,m}) \right)\\
    &= \frac{1}{N} \sum_{n=1}^N h_{\bw}(\btheta_n,\bx_n) 
    - \frac{1}{N} \sum_{n=1}^N \log \left( \frac{1}{M} \sum_{m=1}^M \exp(h_{\bw}( \btheta_{n,m},\bx_n)) \right).
\end{align}

Since in all mentioned methods $\rhat_{\bw}(\bx \mid \btheta)$ is meant to approximate the ratio $\frac{p(\btheta \mid \bx)}{p(\btheta)}$, and estimating the normalizing constant is expensive, a naive alternative to approximate the expected Kullback-Leibler divergence is by plugging the unnormalized distribution $\hat{q}_{\bw}(\btheta \mid \bx) := \rhat_{\bw}(\bx \mid \btheta) p(\btheta)$ into the above formula and using the estimate:
\begin{align}
    \hat{I}_{\bw}(\btheta; \bx) &:= \frac{1}{N} \sum_{n=1}^N  \log \rhat_{\bw}(\bx_n \mid \btheta_n)
    = \frac{1}{N} \sum_{n=1}^N  h_{\bw}(\btheta_n,\bx_n).
    \label{eqn:empirical-mutual-information-by-ratio}
\end{align}
While this is justified for the training objectives for \NREA and \NREC, which encourage a trivial normalizing constant $Z_{\bw}(\bx) \approx 1$ at optimum, the same is not true for \NREB, which leads to an additional non-vanishing, possibly arbitrarily big, bias term:
\begin{align}
    \hat{I}_{\bw}(\btheta; \bx) &\approx I_{\bw}^{(0)}(\btheta; \bx) + C_{\bw}  & (\NREB)
\end{align}

Another way to address the normalizing constant is the use of the Kullback-Leibler divergence that also works for unnormalized distributions $p(\bz)$, $q(\bz)$:
\begin{align}
    \KLD( p(\bz) \Mid q(\bz)) &:= \int  \left( \log\left( \frac{p(\bz)}{q(\bz)}\right) + \frac{q(\bz)}{p(\bz)} -1 \right) \, p(\bz) \, d\bz,
\end{align}
which is always $\ge 0$ with equality if $p(\bz)=q(\bz)$ for $p(\bz)$-almost-all $\bz$.

This gives for the expected Kullback-Leibler divergence between the posterior $p(\btheta \mid \bx)$ and the unnormalized approximate posterior $\hat{q}_{\bw}(\btheta \mid \bx) := \rhat_{\bw}(\bx \mid \btheta) p(\btheta)$:
\begin{align}
  &\qquad \mathbb{E}_{p(\bx)} \left[ \KLD(p(\btheta \mid \bx) \Mid \hat{q}_{\bw}(\btheta \mid \bx))\right] \\
    &= \mathbb{E}_{p(\btheta, \bx)} \left[ 
		\log \frac{p(\btheta \mid \bx)}{\hat{q}_{\bw}(\btheta \mid \bx)} +  \frac{\hat{q}_{\bw}(\btheta \mid \bx)}{p(\btheta \mid \bx)} -1 \right]\\
    &= \mathbb{E}_{p(\btheta, \bx)} \left[ 
		\log \frac{p(\btheta,\bx)}{p(\btheta)p(\bx)} \frac{p(\btheta)}{\hat{q}_{\bw}(\btheta \mid \bx)} +  \frac{\hat{q}_{\bw}(\btheta \mid \bx) p(\bx)}{p(\btheta, \bx)} -1 \right] \\
	&= I(\btheta; \bx) - \mathbb{E}_{p(\btheta, \bx)} \left[ 
		\log\rhat_{\bw}(\bx \mid \btheta) \right] + \int \left( \int \rhat_{\bw}(\bx \mid \btheta) \, p(\btheta)\, d\btheta \right) \, p(\bx)\,d\bx -1 \\
	&= I(\btheta; \bx) - \mathbb{E}_{p(\btheta, \bx)} \left[ 
		\log \rhat_{\bw}(\bx \mid \btheta) \right] + \mathbb{E}_{p(\bx)} \left[ Z_{\bw}(\bx) -1\right].
\end{align}
We see that the normalizing constant $Z_{\bw}(\bx)$ re-appears, but with a different term.
Similar to before the above can be used for another mutual information approximation given by:
\begin{align}
    I_{\bw}^{(1)}(\btheta; \bx) 
    &:= \mathbb{E}_{p(\btheta, \bx)} \left[ \log \rhat_{\bw}(\bx \mid \btheta) \right] - \mathbb{E}_{p(\bx)} \left[ Z_{\bw}(\bx) - 1 \right] \\
	&= \mathbb{E}_{p(\btheta, \bx)} \left[ \log \rhat_{\bw}(\bx \mid \btheta) \right] - \mathbb{E}_{p(\bx)p(\btheta)} \left[ \rhat_{\bw}(\bx \mid \btheta) -1 \right]\\
	&= \mathbb{E}_{p(\btheta, \bx)} \left[ h_{\bw}(\btheta,\bx) \right] - \mathbb{E}_{p(\bx)p(\btheta)} \left[ \exp(h_{\bw}(\btheta,\bx))-1\right],
\end{align}
which can be estimated via Monte-Carlo, again, by sampling i.i.d.\ $\btheta_n,\btheta_{n,m} \sim p(\btheta)$, $\bx_n \sim p(\bx\mid \btheta_n)$, $n=1,\dots,N$, $m=1,\dots,M$ and then computing:
\begin{align}
    \hat{I}_{\bw}^{(1)}(\btheta; \bx) 
    &:= \frac{1}{N} \sum_{n=1}^N \log \rhat_{\bw}(\bx_n \mid \btheta_n) 
    - \frac{1}{N} \frac{1}{M} \sum_{n=1}^N \sum_{m=1}^M \left( \rhat_{\bw}(\bx_n \mid \btheta_{n,m}) -1 \right)\\
    &= \frac{1}{N} \sum_{n=1}^N h_{\bw}(\btheta_n,\bx_n) 
    - \frac{1}{N} \frac{1}{M} \sum_{n=1}^N \sum_{m=1}^M  \left(\exp(h_{\bw}(\btheta_{n,m},\bx_n))-1\right).
\end{align}
However, the estimate has a high variance due to the difficulty in estimating the normalizing constant. Note that since $\log(r) \le r -1$ we always have the inequalities:
\begin{align}
  I(\btheta; \bx) \ge   I_{\bw}^{(0)}(\btheta; \bx) &\ge  I_{\bw}^{(1)}(\btheta; \bx), & \hat{I}_{\bw}^{(0)}(\btheta; \bx) &\ge  \hat{I}_{\bw}^{(1)}(\btheta; \bx),
\end{align}
showing that $I_{\bw}^{(0)}(\btheta; \bx)$ leads to a tighter approximation to the mutual information $I(\btheta; \bx)$ 
than $I_{\bw}^{(1)}(\btheta; \bx)$.

The procedures above require estimating the (log) normalizing constant or the (log) partition function. Estimates of the log normalizing constant are naively biased. Unbiased estimates of the normalizing constant can be quite expensive and may require techniques like Nested Sampling \cite{Skilling2006}. We note however, that the normalizing constant is tractable with rejection-based Monte Carlo on problems with parameters within low dimensional compact regions and, without rigorous justification, we \emph{assume} the bias on the log normalizing constant is small in this case as well. In this circumstance, $\hat{I}_{\bw}^{(0)}(\btheta; \bx)$ may be a reasonable estimate. In more general situations, the looser bound $\hat{I}_{\bw}^{(1)}(\btheta; \bx)$ looks appealing but the ratio can introduce large variance during estimation. This occurs when the posterior is much narrower than the prior, i.e., when data $\bx$ carries a lot of information about $\btheta$.

\clearpage

\paragraph{Bounds on the mutual information}
There is a connection between the training objective of \NREB and a multi-sample lower bound on the mutual information \cite{poole2019variational}, as noted in \citet{Durkan2020}. Contrastive learning has been explored for estimating the mutual information by \citet{van2018representation} also discussed by \citet{belghazi2018mine}. The bounds we define above are also discussed in detail by \citet{poole2019variational}, although we find computing $\hat{I}_{\bw}^{(0)}(\btheta; \bx)$ and $\hat{I}_{\bw}^{(1)}(\btheta; \bx)$ to be tractable within \SBI with a few caveats discussed above: The computation can require many samples when the posterior is extremely narrow, $\hat{I}_{\bw}^{(0)}(\btheta; \bx)$ has an inherent bias, and $\hat{I}_{\bw}^{(1)}(\btheta; \bx)$ can be high-variance.

We attempted to train a ratio estimator by minimizing $\hat{I}_{\bw}^{(1)}(\btheta; \bx)$ directly on fixed data. Our preliminary experiments found that the C2ST was near unity on the SLCP task. We therefore ended our investigation.

However, in a follow up paper which appeared after this one was accepted, we investigated this objective more fully with positive results \cite{miller2023simulation}! The challenge to overcome was that we needed a proposal distribution closer to the posterior than the prior for the method to work.

\paragraph{Numerical estimates of bounds on mutual information}
In remains unclear how to evaluate the performance of \SBI algorithms across model types without access to the ground truth. Computing $\hat{I}_{\bw}^{(0)}(\btheta; \bx)$ or $\hat{I}_{\bw}^{(1)}(\btheta; \bx)$ as a validation loss is applicable to \NREB and \NREC for all $\gamma$ and $K$. Therefore, we investigate estimating this bound on the mutual information for model comparison and as a surrogate for computing the C2ST across several pieces of simulated data. (It is also noteworthy that this bound is applicable to Neural Posterior Estimation, where the likelihood-to-evidence ratio would be approximated by $p_{\bw}(\btheta \mid \bx) / p(\btheta)$ and $p_{\bw}(\btheta \mid \bx)$ represents an approximate posterior density. However, this case is not investigated further in this work.)

\begin{figure}[htb]
\centering
    \begin{subfigure}[b]{0.13\textwidth}
        \includegraphics[width=\textwidth]{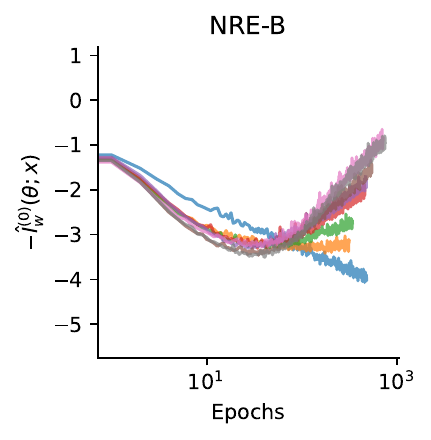}
        \includegraphics[width=\textwidth]{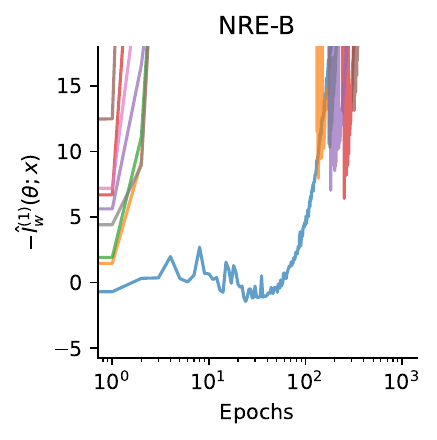}
        \caption{\NREB}
        \end{subfigure}
    \hfill
        \begin{subfigure}[b]{0.84\textwidth}
        \includegraphics[width=\textwidth]{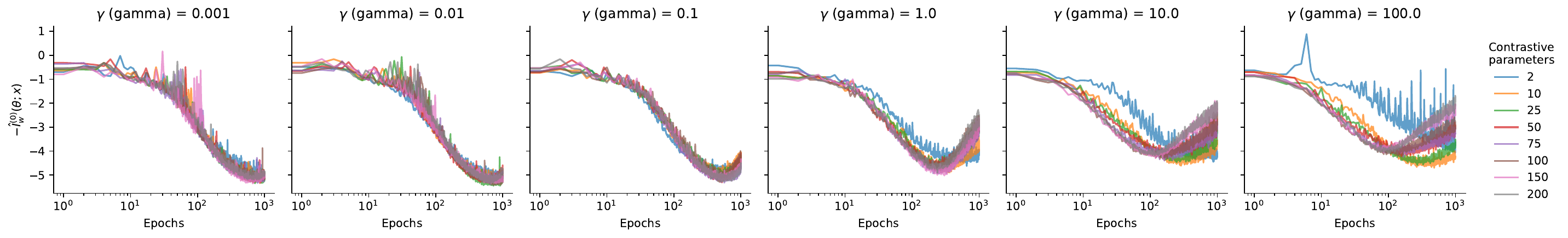}
        \includegraphics[width=\textwidth]{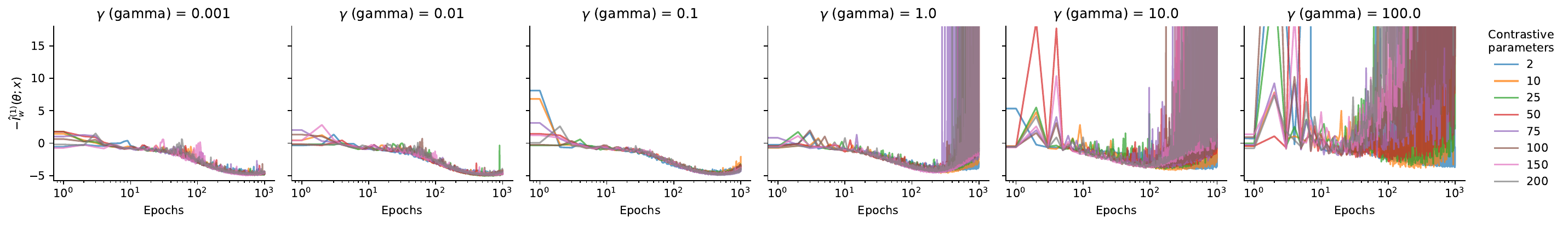}
        \caption{\NREC}
        \end{subfigure}
    \caption{
        A pair of possible metrics, negative bounds on the mutual information $-\hat{I}_{\bw}^{(0)}(\btheta; \bx)$ (top) and $-\hat{I}_{\bw}^{(1)}(\btheta; \bx)$ (bottom), for the SLCP task estimated over the validation set versus training epochs using (a) \NREB and (b) \NREC with various values of $\gamma$ and $K$, a Large NN architecture, and fixed training data.
    }
    \label{fig:mutual-information-0-slcp}
\end{figure}

In the effort to find a model comparison metric that applies when the user does not have access to the ground truth, we ran another set of experiments where we trained ratio estimators using \NREB and \NREC over various $\gamma$ and $K$, then we validated them on held-out data with $-\hat{I}_{\bw}^{(0)}(\btheta; \bx)$ and $-\hat{I}_{\bw}^{(1)}(\btheta; \bx)$ as a validation metric for model comparison. All networks corresponded with the \emph{Large NN} architecture. Just like in the main experiments, we computed the C2ST over the ten different observations from the \SBI benchmark. The results of the training can be seen in Figure~\ref{fig:mutual-information-0-slcp-summary} and in full in Figure~\ref{fig:mutual-information-0-slcp}. 
$-\hat{I}_{\bw}^{(1)}(\btheta; \bx)$ exhibited such high variance that the estimates were hard to compare. 
Visually, $-\hat{I}_{\bw}^{(0)}(\btheta; \bx)$ was more comparable across models than plotting their classification validation loss, see Figure~\ref{fig:slcp-validation-bench}, for both \NREB and \NREC.

\begin{figure}[htb]
    \centering
    \includegraphics[width=0.5\textwidth]{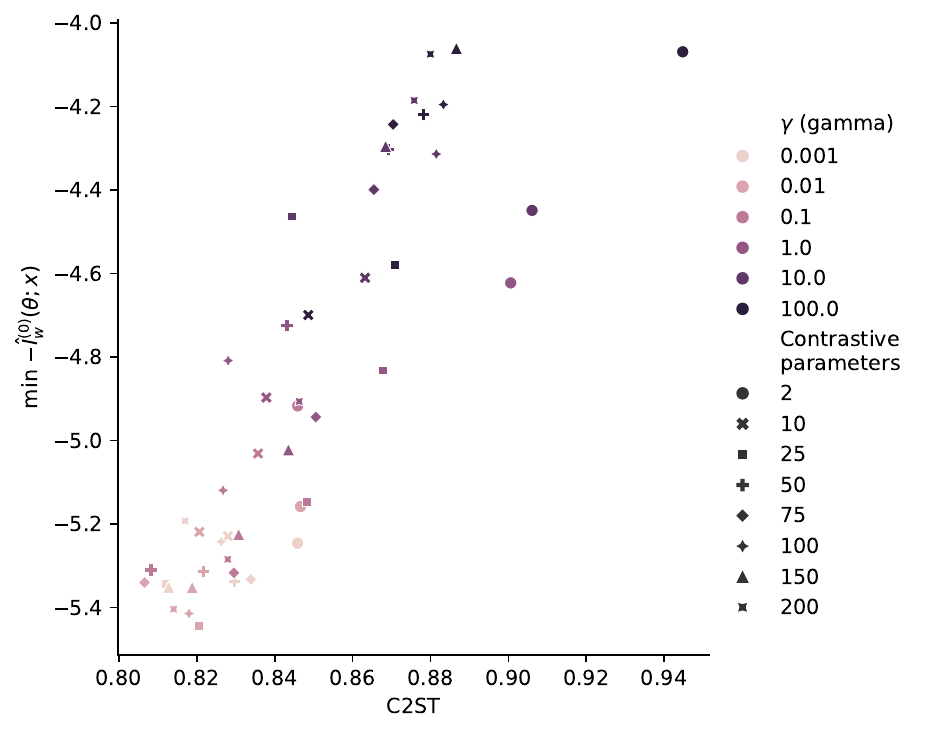}
    \caption{
        A scatter plot of the minimum $-\hat{I}_{\bw}^{(0)}(\btheta; \bx)$ versus the C2ST on the SLCP task with every point corresponding to a different set of values for $\gamma$ and $K$. C2ST scales from 0.5 to 1.0 with 0.5 implying that the classifier could not distinguish the approximate posterior from the ground truth. Just like in the fixed data regime in the main experiment, see Figure~\ref{fig:c2st-specific}, we found that on this task a lower $\gamma$ improved the C2ST. We also find that the $-\hat{I}_{\bw}^{(0)}(\btheta; \bx)$ is correlated with the average C2ST across 10 pieces of data, but $-\hat{I}_{\bw}^{(0)}(\btheta; \bx)$ has the practical advantage that we can bound it without knowing the ground truth posterior. (With the caveat that it has an intrinsic bias when estimated niavely using i.i.d. samples from $p(\btheta, \bx)$.) The C2ST requires being able to sample from the ground truth posterior. This data represents the same set of experiments as in Figure~\ref{fig:mutual-information-0-slcp-summary}, Figure~\ref{fig:mutual-information-0-slcp}, and Figure~\ref{fig:partition-function}.
    }
    \label{fig:mi0-c2st-correlation}
\end{figure}

A correlation plot showing the relationship between $\hat{I}_{\bw}^{(0)}(\btheta; \bx)$ and the C2ST for various $\gamma$ and $K$ on the SLCP task can be found in Figure~\ref{fig:mi0-c2st-correlation}. We find that the two measurements are well correlated. Due to the high variance, we do not compute the same correlation using $-\hat{I}_{\bw}^{(1)}(\btheta; \bx)$. Under circumstances where the variance can be controlled, $-\hat{I}_{\bw}^{(1)}(\btheta; \bx)$, which does not require access to the ground truth posterior, may be able to replace computing the C2ST across several pieces of data, which does require access to the ground truth posterior, but further investigation is necessary.

\clearpage

\begin{figure}[hb]
\centering
    \begin{subfigure}[b]{0.13\textwidth}
        \includegraphics[width=\textwidth]{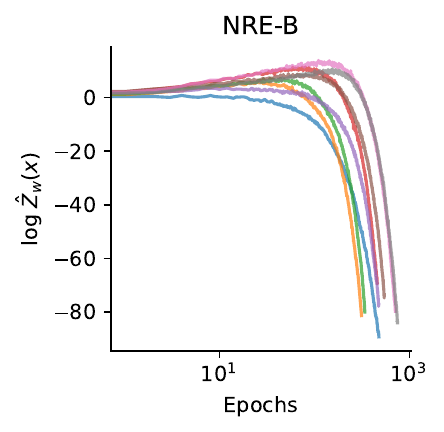}
        \caption{\NREB}
        \end{subfigure}
    \hfill
        \begin{subfigure}[b]{0.84\textwidth}
        \includegraphics[width=\textwidth]{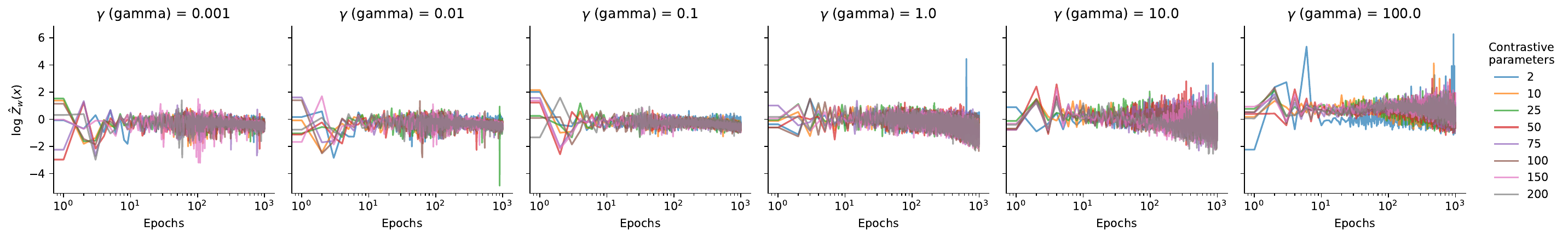}
        \caption{\NREC}
        \end{subfigure}
    \caption{
        Monte Carlo estimate of the partition function $\hat{Z}_{\bw}(\bx)$ for (a) \NREB and (b) \NREC with various values for $\gamma$ and $K$ on the SLCP inference problem. We transformed the estimates using the $\log$ function to show the huge dynamic range caused by \NREB. $\hat{Z}_{\bw}(\bx)$ is completely unconstrained with \NREB; however, it is constrained with \NREC. The strength of encouragement the partition function towards unity depends on the hyperparameters for \NREC.
        This represents the same set of experiments as in Figure~\ref{fig:mutual-information-0-slcp-summary}, Figure~\ref{fig:mutual-information-0-slcp-summary} and Figure~\ref{fig:mi0-c2st-correlation}.
    }
    \label{fig:partition-function}
\end{figure}

\paragraph{Numerical estimates of the partition function}
In addition to bounds on the mutual information, we computed a Monte Carlo estimate of the partition function, $Z_{\bw}(\bx)$, based on data from the validation set at every epoch during training. The value of the estimated partition function as a function of epoch is shown in Figure~\ref{fig:partition-function} for this set of runs on the SLCP task. The estimated $Z_{\bw}(\bx)$, based on the ratio from \NREB, is completely unconstrained and diverges towards $-\infty$ as epochs increase. \NREC successfully encourages the partition function to remain ``near'' unity, although both $\gamma$ and $K$ affect the strength of the encouragement.

This result is connected to the reliability of the importance sampling diagnostic, see Section~\ref{sec:diagnostic}. If the partition function is not near unity, then the estimated likelihood-to-evidence ratio does not cancel with the evidence and the diagnostic will behave like \NREB, i.e., it becomes possible to produce accurate, albeit unnormalized, posteriors while failing the diagnostic.

\end{document}